\documentclass{article}

\usepackage[preprint]{neurips_2026}

\usepackage[utf8]{inputenc}
\usepackage[T1]{fontenc}
\usepackage{microtype}
\usepackage{hyperref}
\usepackage{url}
\usepackage{booktabs}
\usepackage{amsfonts}
\usepackage{nicefrac}
\usepackage[table]{xcolor}
\usepackage{graphicx}
\usepackage{amsmath}
\usepackage{cleveref}
\usepackage[ruled,vlined]{algorithm2e}
\usepackage{multirow}
\usepackage{threeparttable}
\usepackage{float}
\usepackage{wrapfig}
\usepackage{listings}
\usepackage[most]{tcolorbox}

\definecolor{darkblue}{rgb}{0, 0, 0.5}
\definecolor{Gray}{gray}{0.9}
\hypersetup{colorlinks=true, citecolor=darkblue, linkcolor=darkblue, urlcolor=darkblue}
\newtcblisting{promptbox}[1]{
  breakable,
  enhanced,
  colback=black!2,
  colframe=black!20,
  colbacktitle=black!8,
  coltitle=black,
  boxrule=0.4pt,
  arc=1.5mm,
  left=1mm,
  right=1mm,
  top=1mm,
  bottom=1mm,
  title={#1},
  fonttitle=\bfseries,
  listing only,
  listing options={
    basicstyle=\ttfamily\footnotesize,
    breaklines=true,
    columns=fullflexible,
    keepspaces=true
  }
}
\title{AdaTurn: Budget-Aware Test-Time Scaling for \\ Active Visual Perception Agents}

\author{
\textbf{Susan Liang$^1$}
\quad
\textbf{Chao Huang$^1$}
\quad
\textbf{Filippos Bellos$^2$}
\\
\textbf{Jing Bi$^1$}
\quad
\textbf{ Jason J Corso$^2$}
\quad
\textbf{Chenliang Xu$^1$}
\\
$^1$University of Rochester \quad $^2$Univeristy of Michigan
}

\begin{document}

\maketitle

\begin{abstract}
Active visual agents solve fine-grained image tasks by interleaving reasoning with image-grounding actions across multiple turns. However, deployment-time rollout budgets are rarely fixed: some requests permit long rollouts, while others require the agent to act under a tight turn limit. Existing methods train the policy as if the rollout budget were hidden, so when the available budget is smaller than the trajectory the agent prefers, the interaction is often truncated before any valid answer is produced; we term this failure \emph{catastrophic truncation}. To overcome this challenge, we present AdaTurn, a budget-aware framework that conditions the agent on the allowed number of turns and explicitly trains the boundary behavior induced by the budget. Our key component, Forced-Answer DAPO (FA-DAPO), converts the over-budget event from a masked or penalized failure into a trainable final-decision step, teaching the model to synthesize partial evidence when further tool use is no longer possible. We further randomize rollout budgets during both training and inference and introduce a load-balanced scheduler that makes such operations practical. AdaTurn substantially improves low-budget accuracy, for example raising VisualProbe-Medium from 36.7\% to 47.6\% at four turns, while preserving strong scaling at larger budgets and transferring effectively to multiple backbones and general multimodal benchmarks.
\end{abstract}

\begin{figure}[H]
  \centering
  \includegraphics[width=\linewidth]{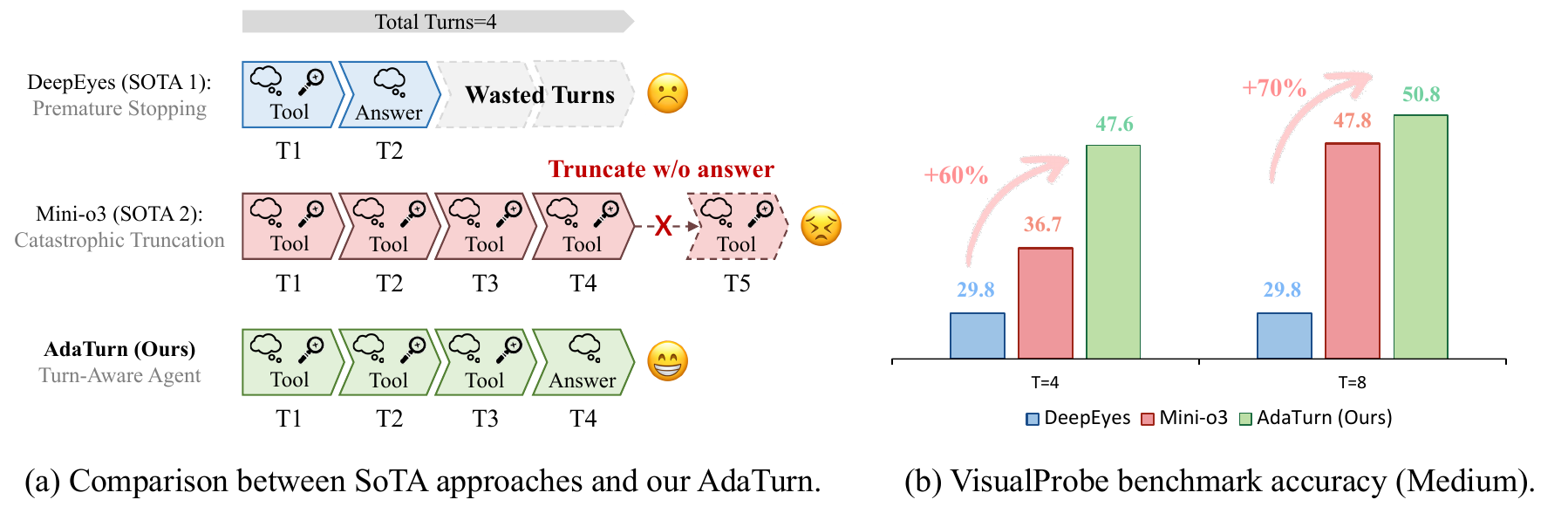}
  \caption{Turn-aware reasoning under limited rollout budgets. Existing visual agents can stop too early when extra turns remain or suffer catastrophic truncation when the allocated budget is smaller than the rollout they implicitly expect, yielding no valid answer. AdaTurn conditions the agent on the turn budget and teaches it to switch from tool use to answer generation before the budget is exhausted, which leads to much stronger low-budget performance while preserving high-budget gains.}
  \label{fig:teaser}
\end{figure}

\section{Introduction}
\label{sec:intro}

Multimodal language models are increasingly used not only as one-shot image predictors~\citep{liu2023llava,xu2024llavauhdlmmperceivingaspect,bai2025qwen25vl,chen2024expanding}, but also as active visual agents that interleave reasoning with image-grounding actions across multiple turns~\citep{wu2024vstar,wang2024dc2,zhang2025chainoffocus,shen2024zoomeye,zheng2025deepeyes,su2025pixel,lai2025minio3}. Instead of answering from a single encoded image, the model can inspect local regions, gather new observations, and refine its decision over multiple interaction turns.

This shift has made test-time scaling newly relevant for visual reasoning~\citep{snell2024scaling,openai2024o1,agarwal2025art,wang2025scaling}. Recent systems such as DeepEyes~\citep{zheng2025deepeyes}, Pixel Reasoner~\citep{su2025pixel}, Chain-of-Focus~\citep{zhang2025chainoffocus}, and Mini-o3~\citep{lai2025minio3} show that ``thinking with images'' can outperform standard one-pass prediction on high-resolution and visually cluttered tasks. When the agent is allowed more turns, it can search more carefully, reject distractors, and verify a tentative answer before committing.

Yet this success hides an important deployment mismatch. In practical settings, the rollout budget is an external resource constraint rather than a fixed property of the model~\citep{wen2025budgetthinker,nogueira2025certainty}. One request may permit only four turns because of latency requirements, while another may permit thirty-two. Existing visual agents are typically trained for a fixed rollout regime rather than with the deployment-time budget as an explicit input~\citep{zheng2025deepeyes,su2025pixel,lai2025minio3}. As a result, when the available budget is smaller than the policy prefers, the agent may continue requesting tools until the system terminates the rollout without any valid answer. We call this failure mode \emph{catastrophic truncation}. For example, Mini-o3's performance on VisualProbe-Hard benchmark reduces from 48.0\% to 26.5\% after we limit the maximum rollout turns to 4. The same budget-blindness also wastes computation at the opposite extreme: when extra turns are available, the agent may stop as soon as it finds a plausible answer instead of using the remaining budget for verification.

AdaTurn addresses this mismatch by making the rollout budget part of the learning problem. The central idea is simple: a capable visual agent should not learn a single rollout style, but a family of behaviors calibrated to the available budget. This perspective yields three design requirements. First, the policy must be conditioned on the maximum allowed turns, so that short-budget and long-budget behavior can differ systematically. Second, the model must receive learning signal at the budget boundary, because this is precisely where standard training breaks down. Third, if rollout budgets vary during training and inference, the underlying infrastructure must also become budget-aware; otherwise the slowest worker dominates training throughput.

We satisfy these requirements with AdaTurn, an active visual perception agent with dynamic rollout depth. Besides the budget-conditioned visual agent, we also introduce \textbf{Forced-Answer DAPO (FA-DAPO)}, a reinforcement learning strategy built on DAPO~\citep{yu2025dapo}, a more robust variant of GRPO~\citep{shao2024deepseekmath}, that converts an over-budget tool request into a trainable final-answer decision. Instead of penalizing or masking the boundary case, AdaTurn teaches the policy how to synthesize the evidence collected so far when no further tool use is possible. We further train with dynamic rollout budgets and pair this with a load-balanced scheduler that keeps worker utilization stable despite heterogeneous rollout lengths.

The resulting system is effective across different rollout budgets and benchmarks. At four turns, AdaTurn improves VisualProbe-Medium~\citep{lai2025minio3} from 36.7\% to 47.6\% and MME-RealWorld~\citep{zhang2024mmerealworld} from 50.1\% to 64.0\% over Mini-o3~\citep{lai2025minio3}. Figure~\ref{fig:sota} shows that these gains are concentrated where deployment constraints matter most, namely the low-budget regime, while the model remains strong as the budget increases. We also find that the approach transfers to Qwen3-VL 4B and 8B backbones~\citep{bai2025qwen3} and preserves broad multimodal ability on OCR and reasoning benchmarks.

In summary, our contributions are:
\begin{itemize}
    \item We formulate budget-aware active visual perception and identify catastrophic truncation as a central failure mode of deployment-constrained visual agents.
    \item We propose AdaTurn, a budget-conditioned visual agent, and Forced-Answer DAPO (FA-DAPO), which explicitly trains the final-answer decision at the budget boundary.
    \item We introduce a load-balanced scheduler that reduces rollout imbalance during both training and inference and improves rollout throughput by $1.34\times$.
    \item We demonstrate that AdaTurn delivers large gains in the low-budget regime, remains strong at larger budgets, transfers to multiple backbones, and preserves general multimodal capability.
\end{itemize}

\section{Related Work}
\label{sec:related}

\subsection{High-Resolution Image Understanding}
Early vision-language models such as LLaVA~\citep{liu2023llava} relied on fixed low-resolution inputs, which discard the small details that often determine success on fine-grained visual tasks. More recent high-resolution models, including LLaVA-UHD~\citep{xu2024llavauhdlmmperceivingaspect}, Qwen2.5-VL~\citep{bai2025qwen25vl}, and InternVL 2.5~\citep{chen2024expanding}, improve image encoding through dynamic tiling, image slicing, or stronger visual backbones. However, these approaches still perform inference largely in one pass, so they cannot adaptively allocate attention to the regions that become relevant during reasoning. This limitation has motivated active perception methods such as SEAL~\citep{wu2024vstar}, DC$^{2}$~\citep{wang2024dc2}, Chain-of-Focus~\citep{zhang2025chainoffocus}, and ZoomEye~\citep{shen2024zoomeye}, which explicitly inspect the image over multiple steps. AdaTurn operates in this active-perception regime, but focuses on a different question: how such agents should behave when the available rollout budget itself varies across requests.

\subsection{Agentic Visual Perception}
Recent work has shown that multi-turn visual search can be learned directly within a language-model-style action loop. DeepEyes~\citep{zheng2025deepeyes} demonstrates that image-grounded tool use can emerge from reinforcement learning alone, while DeepEyesV2~\citep{hong2025deepeyesv2}, VISTA-R1~\citep{lu2025vista}, and Agent0-VL~\citep{liu2025agent0vl} extend the tool space or the training environment. Mini-o3~\citep{lai2025minio3} is particularly relevant because it shows that visual agents can benefit substantially from longer rollouts and richer interaction patterns. Our work differs in focus --- we do not primarily ask how to obtain longer trajectories; we ask how the same agent should behave when the permitted trajectory length changes at deployment time. AdaTurn treats the rollout budget as an explicit conditioning variable and optimizes the final-turn decision directly, which addresses a failure mode that prior visual agents largely leave unresolved.

\subsection{Reinforcement Learning and Test-Time Scaling for VLMs}
Group Relative Policy Optimization (GRPO)~\cite{shao2024deepseekmath} has become a standard recipe for training reasoning models, beginning with DeepSeek-R1~\citep{guo2025deepseek} and extending rapidly to multimodal settings such as VLM-R1~\citep{shen2025vlmr1}, R1-VL~\citep{zhang2025r1vl}, and Visionary-R1~\citep{zhang2025visionaryr1}. Beyond one-shot reasoning, recent sequence-level reinforcement learning formulations have also begun to study multi-turn tool-integrated reasoning more directly~\citep{simpletir,gspo}. In parallel, test-time scaling has emerged as a dominant view of reasoning systems: additional inference-time computation often improves performance, but only when the model knows how to use that extra budget effectively~\citep{snell2024scaling,openai2024o1,agarwal2025art,wang2025scaling}. Prior budget-aware methods have mainly focused on token-level control within a single generation, for example via budget tokens or confidence-based early stopping~\citep{wen2025budgetthinker,nogueira2025certainty}. AdaTurn addresses a different setting, namely multi-turn agents whose action space changes at the final turn because tool execution is no longer available. Our contribution is therefore not only budget conditioning, but boundary-aware reinforcement learning and the systems support needed to train it efficiently.

\section{Method}
\label{sec:method}
AdaTurn is built around a simple premise: for an active visual agent, the rollout budget is not a nuisance variable but part of the task specification. The policy should therefore depend not only on the image and the question, but also on how many interaction turns are available. We first formalize this budget-aware setting in Section~\ref{sec:formulation}. We then present Forced-Answer DAPO (FA-DAPO) in Section~\ref{sec:fa-grpo}, which supplies the missing learning signal at the final-turn boundary. Finally, Section~\ref{sec:dynamic-load-balance} describes dynamic-budget training and the load-balanced rollout scheduler that makes such training efficient in practice. Figure~\ref{fig:framework} gives an overview.

\subsection{Problem Formulation and AdaTurn Agent Loop}
\label{sec:formulation}
\begin{figure}[t]
  \centering
  \includegraphics[width=\linewidth]{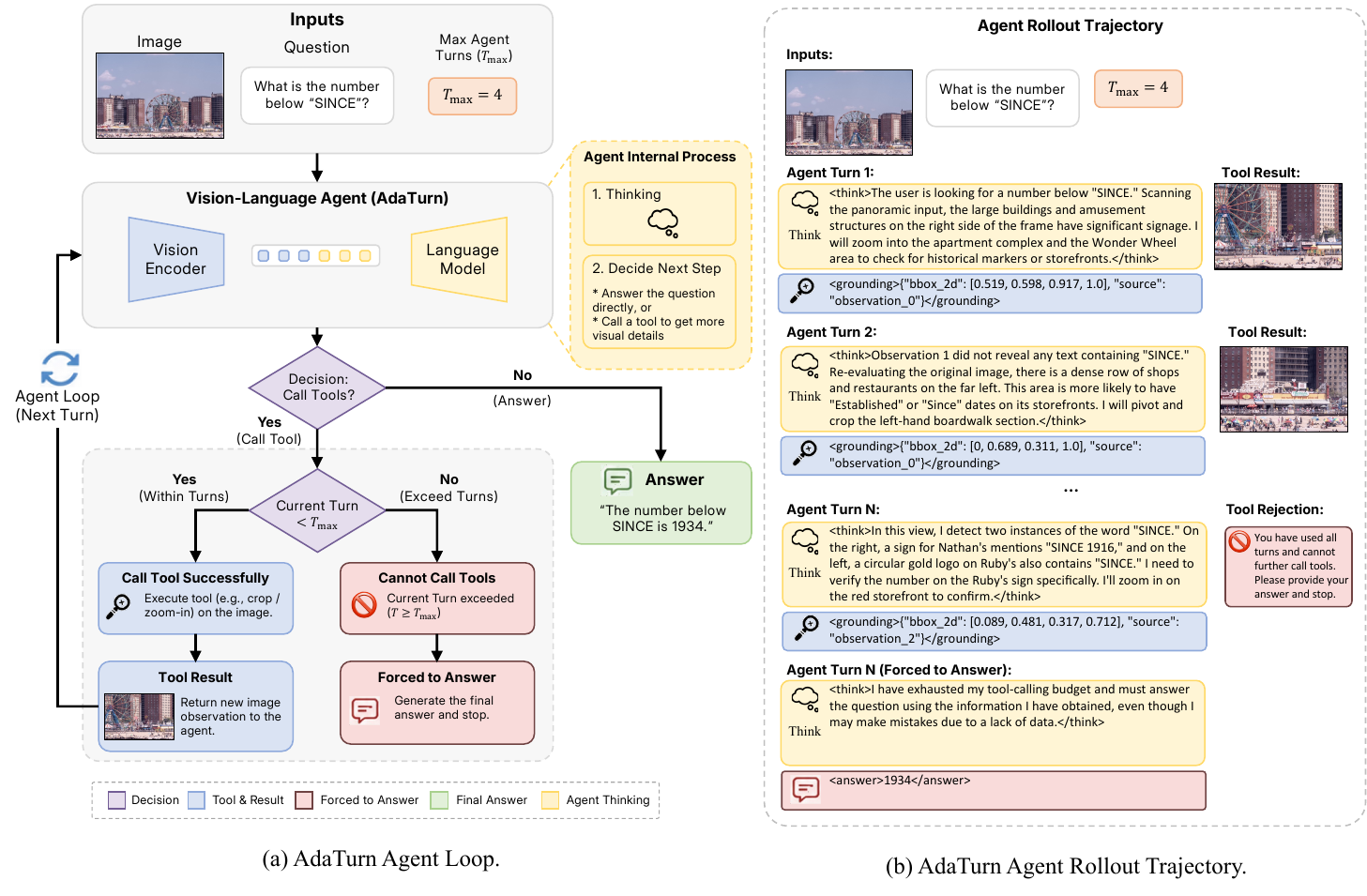}
  \vspace{-2mm}
  \caption{Overview of AdaTurn. The left panel shows the turn-aware agent loop: the model receives the image, the user question, and the maximum rollout budget $T_\mathrm{max}$, then decides at each turn whether to call a visual tool or emit the final answer. If a tool request would exceed the budget, AdaTurn blocks the tool call and forces the model to answer using the evidence already collected. The right panel illustrates a concrete rollout in which the agent gathers several observations and then produces a budget-aware final answer at the last turn.}
  \label{fig:framework}
\end{figure}

Let $Q$ denote the user question, $I$ the input image, $T_\mathrm{max}$ the maximum rollout budget, and $g$ the ground-truth answer. AdaTurn solves a budget-aware decision problem with input
\begin{equation}
x = (Q, I, T_\mathrm{max}).
\label{eq:task}
\end{equation}
At turn $t$, the agent conditions on the interaction history
\begin{equation}
h_t = \left(Q, I, T_\mathrm{max}, (y_1, o_1), \ldots, (y_{t-1}, o_{t-1})\right),
\label{eq:history}
\end{equation}
where $y_t$ is the token sequence generated at turn $t$ and $o_t$ is the tool observation returned by the environment. At each turn, the policy $\pi_\theta$ emits either a tool-calling sequence $c_t\in \mathcal{C}$ or a final-answer sequence $a_t \in \mathcal{A}$:
\begin{equation}
y_t \sim \pi_\theta(\cdot \mid h_t), \qquad y_t \in \mathcal{C} \cup \mathcal{A}.
\label{eq:policy}
\end{equation}
If $y_t \in \mathcal{C}$ and $t < T_\mathrm{max}$, the tool executor $\mathcal{E}$ returns a new observation
\begin{equation}
o_t = \mathcal{E}(I, y_t),
\label{eq:tool}
\end{equation}
which becomes part of the next-turn history. If $y_t \in \mathcal{A}$, the rollout terminates and the answer is scored against $g$. The key structural constraint appears at the last turn:
\begin{equation}
y_t \in \mathcal{Y}_t =
\begin{cases}
\mathcal{C} \cup \mathcal{A}, & t < T_\mathrm{max},\\
\mathcal{A}, & t = T_\mathrm{max}.
\end{cases}
\label{eq:feasible-actions}
\end{equation}
Before the final turn, the agent may either gather more evidence or answer. At the final turn, only answer actions are executable. Catastrophic truncation arises when the learned policy still prefers a tool action at this boundary.

Figure~\ref{fig:framework} illustrates the resulting agent loop. The left panel shows the budget-conditioned control flow: the model receives the image, the question, and the maximum budget $T_\mathrm{max}$, then repeatedly decides whether to spend another turn on grounding or to terminate with an answer. This makes the budget visible throughout the rollout rather than leaving it as an implicit system-side constraint. The right panel shows a concrete trajectory in which successive observations narrow the search space until the policy emits a final answer at the budget boundary. The important point is not merely that AdaTurn stops at turn $T_\mathrm{max}$, but that it is trained to make this boundary decision intelligently. Under large budgets, the policy can spend additional turns on verification; under small budgets, it learns to compress the search and commit using partial evidence.

\subsection{Forced-Answer DAPO}
\label{sec:fa-grpo}
\begin{figure}[t]
   \centering
   \includegraphics[width=\linewidth]{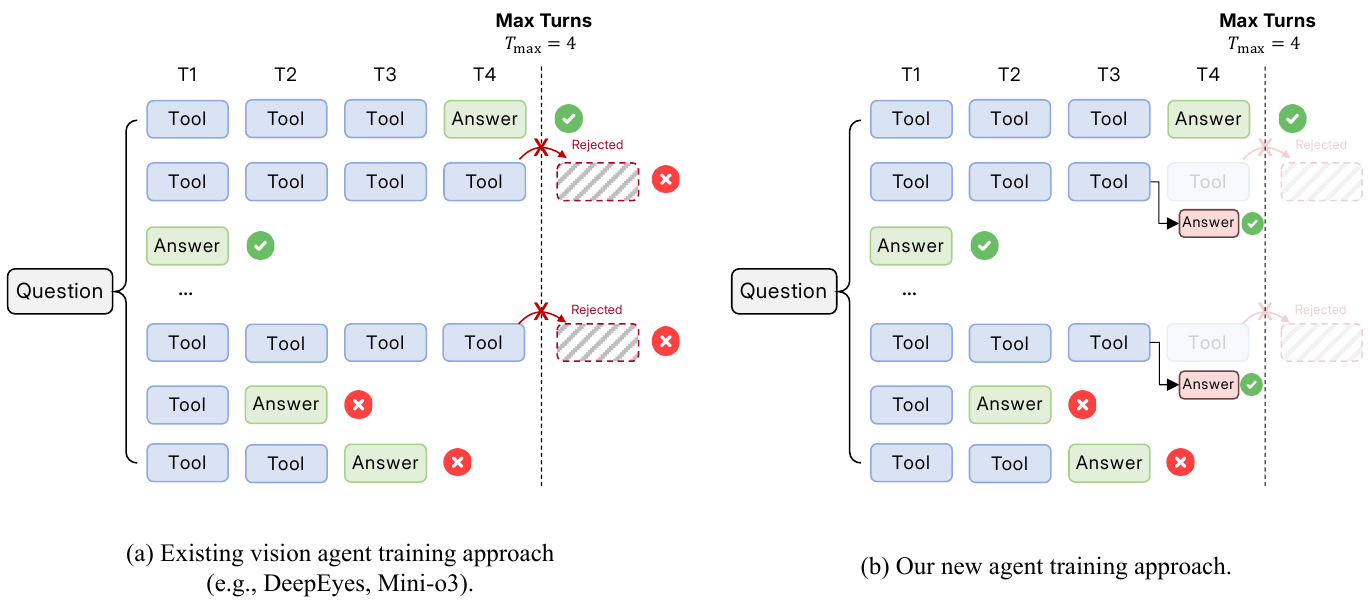}
   \vspace{-2mm}
   \caption{Training behavior at the budget boundary. Prior methods either treat over-budget rollouts as wrong and penalize them, or mask them and discard the learning signal. Forced-Answer DAPO instead rolls back to the last valid prefix and explicitly trains the model to produce a final answer when the next tool request would exceed the budget.}
   \label{fig:grpo}
\end{figure}

We adopt Decoupled Clip and Dynamic Sampling Policy Optimization (DAPO)~\citep{yu2025dapo}, a more robust variant of GRPO~\citep{shao2024deepseekmath,guo2025deepseek,shen2025vlmr1,zhang2025r1vl}, as the reinforcement learning algorithm. For each input $x$, we sample a group of $G$ rollouts $\{Y^{(i)}\}_{i=1}^G$ from the current policy, score them with rewards $\{R^{(i)}\}_{i=1}^G$, and compute normalized advantages
\begin{equation}
\hat{A}^{(i)} = \frac{R^{(i)} - \frac{1}{G}\sum_{j=1}^{G} R^{(j)}}{\mathrm{Std}(\{R^{(j)}\}_{j=1}^{G}) + \delta},
\label{eq:adv}
\end{equation}
where $Y^{(i)}=[y^{(i)}_1, \dots, y^{(i)}_{T_\mathrm{max}}]$ is the $i$-th rollout and $\delta$ is a small constant for numerical stability. DAPO then optimizes the token-normalized clipped objective:
\begin{equation}
\mathcal{L}_{\text{DAPO}}(\theta) =
- \mathbb{E}\left[
\frac{1}{\sum_{i=1}^{G} |Y^{(i)}|}\sum_{i=1}^{G}\sum_{k}
\min\!\left(
\rho_{i,k}\hat{A}_{i,k},
\mathrm{clip}(\rho_{i,k}, 1-\epsilon_{\mathrm{low}}, 1+\epsilon_{\mathrm{high}})\hat{A}_{i,k}
\right)
\right]
\label{eq:grpo}
\end{equation}
where $\hat{A}_{i,k}=\hat{A}^{(i)}$ for each token $k$ in rollout $Y^{(i)}$, and the token-wise importance ratio is
\begin{equation}
\rho_{i,k} =
\frac{\pi_\theta\!\left(y^{(i)}_{k}\mid x, y^{(i)}_{<k}\right)}
{\pi_{\theta_{\mathrm{old}}}\!\left(y^{(i)}_{k}\mid x, y^{(i)}_{<k}\right)}.
\label{eq:ratio}
\end{equation}
With different clip values ($\epsilon_{\mathrm{low}}$ and $\epsilon_{\mathrm{high}}$ for positive and negative advantages, respectively) and token-wise normalization, DAPO can stabilize training while still allowing the policy to learn from outlier samples with large advantages.

The central difficulty is that the final-turn boundary is qualitatively different from the rest of the rollout. At turns $t<T_\mathrm{max}$, the policy decides between continuing to gather evidence and terminating. At turn $t=T_\mathrm{max}$, that choice collapses: tool calls are no longer feasible. Figure~\ref{fig:grpo} shows that existing strategies do not optimize this regime well. DeepEyes~\citep{zheng2025deepeyes} treats the over-budget case as a wrong rollout, which discourages long-horizon exploration and effectively ties the policy to the training budget. Mini-o3~\citep{lai2025minio3} avoids this penalty by masking truncated cases, but this also removes the learning signal precisely where budget-awareness matters most. In both cases, the boundary behavior is under-trained.

We address this with Forced-Answer DAPO (FA-DAPO). Suppose a rollout reaches turn $T_\mathrm{max}$ and the policy still prefers a tool request. Instead of treating the sample as either incorrect or irrelevant, we roll back to the last executable prefix $h_{T_\mathrm{max}}^{-}$ and append a control instruction stating that the tool budget has been exhausted and the model must provide its best final answer. Formally, we transform each sampled rollout as
\begin{equation}
\tilde{Y}^{(i)} =
\begin{cases}
Y^{(i)}, & \text{if } Y^{(i)} \text{ emits an answer within } T_\mathrm{max},\\
\left(h_{T_\mathrm{max}}^{-(i)}, \tilde{a}_{T_\mathrm{max}}^{(i)}\right), & \text{if } y_{T_\mathrm{max}}^{(i)} \in \mathcal{C},
\end{cases}
\label{eq:forced-answer}
\end{equation}
where $\tilde{a}_{T_\mathrm{max}}^{(i)} \sim \pi_\theta(\cdot \mid h_{T_\mathrm{max}}^{-(i)}, b_{T_\mathrm{max}})$ and $b_{T_\mathrm{max}}$ is the budget-exhausted control message. We then score the resulting answer with
\begin{equation}
R^{(i)} = \lambda_{\mathrm{acc}}\, r_{\mathrm{acc}}(\tilde{Y}^{(i)} , g) + \lambda_{\mathrm{fmt}}\, r_{\mathrm{fmt}}(\tilde{Y}^{(i)} ).
\label{eq:reward}
\end{equation}
Conceptually, FA-DAPO converts catastrophic truncation from an artifact of the rollout engine into a supervised decision problem for the policy. This design enables AdaTurn to operate reliably under tight budgets while still benefiting from additional turns when they are available. We examine several alternative training variants, including masked-prefix optimization, final-turn-only reinforcement learning, and format-weighted rewards, in Section~\ref{sec:ablation}.

\begin{figure}[t]
  \centering
  \includegraphics[width=0.8\linewidth]{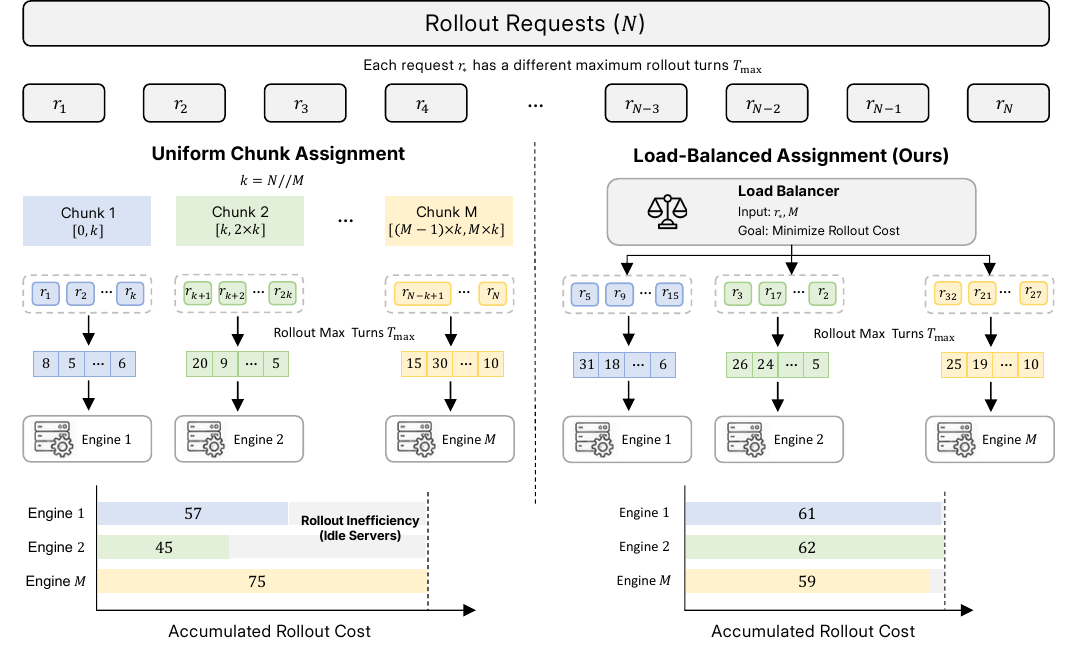}
  \caption{Load-balanced rollout assignment. Dynamic rollout budgets create large differences in per-sample rollout cost. Instead of uniformly splitting requests across engines, we sort requests by assigned budget and greedily place each one on the engine with the current lowest accumulated load, which keeps the rollout workers much better balanced.}
  \label{fig:load_balance_logic}
\end{figure}

\subsection{Dynamic Rollout Budgets and Load-Balanced Scheduling}
\label{sec:dynamic-load-balance}

The maximum rollout budget is not fixed during either training or inference. During training, we sample $T_\mathrm{max}$ from a predefined set of candidate budgets and inject it into the prompt as part of the agent state. The same task therefore appears under both short and long budgets, forcing the policy to learn a calibrated trade-off between further search and immediate commitment. This same mechanism carries over directly to inference: practitioners can set $T_\mathrm{max}$ according to their latency or compute budget without switching models or retraining the policy.

Dynamic budgets, however, create a systems bottleneck. In multi-engine rollout generation, batch latency is determined by the slowest engine. If long-budget samples are assigned unevenly, fast workers idle while one overloaded worker processes a set of overlong trajectories. Figure~\ref{fig:load_balance_logic} shows our solution. We assign rollout jobs by sorting requests according to their assigned budget and greedily dispatching each request to the engine with the current lowest accumulated load, which keeps the rollout cost more balanced across workers.
Appendix~\ref{sec:appendix-load-balance} provides the full pseudocode. Each request $r_n$ carries an assigned rollout budget $T_n$. After sorting requests by $T_n$ in descending order, we greedily assign each request to the engine with the current lowest accumulated load. We quantify rollout imbalance as the gap between the largest and smallest accumulated rollout budgets across engines after assignment.

\begin{figure}[t]
  \centering
  \includegraphics[width=0.9\linewidth]{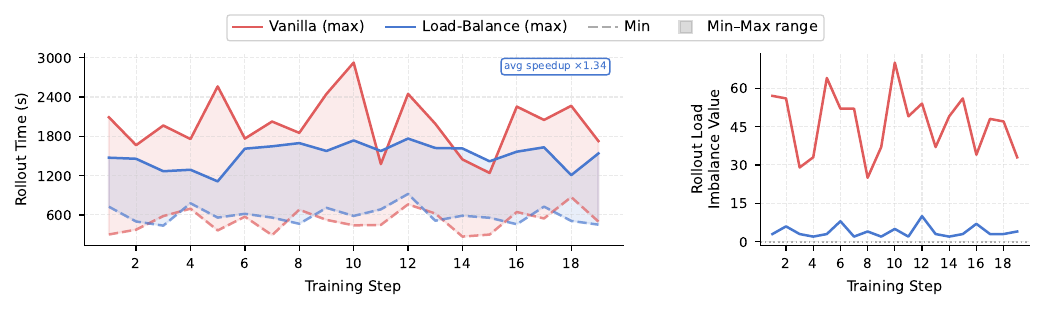}
  \vspace{-2mm}
  \caption{Effect of load-balanced rollout scheduling. The left plot reports the minimum, maximum, and min-max rollout time range across engines, showing that the scheduler reduces the maximum engine time and yields an average $1.34\times$ speedup. The right plot reports the gap between the largest and smallest accumulated rollout budgets across engines, which is consistently smaller with load balancing.}
  \label{fig:load_balance_result}
\end{figure}

Figure~\ref{fig:load_balance_result} reports the effect on real training runs. The scheduler consistently reduces both the maximum rollout time and the cross-engine load imbalance value calculated as the gap between the largest and smallest accumulated rollout budgets across engines. This leads to an average $1.34\times$ speedup in rollout generation, which is critical for dynamic-budget training and inference.

\begin{table}[t]
  \centering
  \caption{Quantitative comparison on visual perception benchmarks. AdaTurn performs favorably in the low-budget regime and remains competitive at 32 turns while using the same 7B backbone scale as prior open-source agent baselines. For VisualProbe and V\textsuperscript{*} Bench, we report Avg@32 to reduce variance caused by randomness. We report Avg@8 and Avg@1 for HR-Bench and MME-RealWorld, respectively.}
  \label{tab:main}
  \vspace{0.1cm}
  \tabcolsep=0.25cm
  {
      \begin{threeparttable}
      \resizebox{\linewidth}{!}{
      \begin{tabular}{l|ccc|c|cc|c}
          \toprule
          \multirow{2}{*}{Model}
          & \multicolumn{3}{c|}{VisualProbe}
          & \multirow{2}{*}{V*}
          & \multicolumn{2}{c|}{HR-Bench} & \multirow{2}{*}{MME-RealWorld} \\
          
          \specialrule{0em}{0pt}{1pt}
          \cline{2-4}
          \cline{6-7}
          \specialrule{0em}{0pt}{1pt}
          
          & hard & medium & easy &  & 4K & 8K \\
          \midrule
          GPT-4o~\citep{gpt4o} & 11.2 & 15.4 & 47.5 & 65.2 & 62.0 & 58.3 & 45.2 \\

          \specialrule{0em}{0pt}{1pt}
          \hline
          \specialrule{0em}{0pt}{1pt}

          LLaVA-OneVision~\citep{li2024llava} & 13.4 & 12.5 & 36.2 & 70.9 & 61.2 & 54.0 & 57.4 \\
          Qwen2.5-VL-Instruct~\citep{bai2025qwen25vl} & 23.9 & 26.0 & 39.1 & 75.5 & 68.2 & 62.7 & 57.3 \\
          \specialrule{0em}{0pt}{1pt}
          \hline
          \specialrule{0em}{0pt}{1pt}
          SEAL~\citep{wu2024v} & - & - & - & 75.4 & - & - & - \\
          DyFo~\citep{li2025dyfo} & - & - & - & 81.2 & - & - & - \\
          Chain-of-Focus~\citep{zhang2025chain} & - & - & - & 88.0 & - & - & - \\
          Pixel Reasoner~\citep{su2025pixel} & 28.8 & 29.6 & 58.4 & 86.3 & 74.0 & 66.9 & 64.4 \\
          DeepEyes~\citep{zheng2025deepeyes} (6 turns) & 35.1 & 29.8 & 60.1 & 83.3 & 73.2 & 69.5 & 64.0 \\
          Mini-o3~\citep{lai2025minio3} (4 turns) & 26.5 & 36.7 & 42.4 & 79.2 & 68.9 & 63.2 & 50.1 \\
          Mini-o3~\citep{lai2025minio3} (32 turns) & 48.0 & 50.4 & \textbf{67.0} & \textbf{88.2} & 77.5 & 73.3 & 65.5 \\
          \rowcolor{Gray}Ours (4 turns) & \textbf{39.0} & \textbf{47.6} & \textbf{61.2} & \textbf{85.2} & \textbf{75.7} & \textbf{71.1} & \textbf{64.0}\\
          \rowcolor{Gray}Ours (32 turns) & \textbf{48.1} & \textbf{51.0} & 66.7 & 86.9 & \textbf{77.5} & \textbf{73.4} & \textbf{66.3}\\
          \bottomrule
      \end{tabular}
      }
      \end{threeparttable}
  }
\end{table}

\section{Experiments}
\label{sec:experiments}

\subsection{Experimental Details}
\label{sec:exp-setup}
\noindent\textbf{Datasets.} We evaluate AdaTurn on VisualProbe~\citep{lai2025minio3} and V\textsuperscript{*} Bench~\citep{wu2024v} for agentic visual search, HR-Bench~\citep{wang2024dc2} at 4K and 8K for high-resolution perception, and MME-RealWorld~\citep{zhang2024mmerealworld} for real-world multimodal reasoning. To assess whether the resulting policy remains broadly useful beyond the main visual-search setting, we additionally report OCRBench~\citep{liu2024ocrbench}, ChartQA~\citep{masry2022chartqa}, DocVQA~\citep{mathew2021docvqa}, MathVista~\citep{lu2023mathvista}, ScienceQA~\citep{lu2022learn}, and CV-Bench~\citep{tong2024cambrian}.

\noindent\textbf{Baselines.} We compare against closed and open-source multimodal models, including GPT-4o~\citep{gpt4o}, LLaVA-OneVision~\citep{li2024llava}, Qwen2.5-VL-7B-Instruct~\citep{bai2025qwen25vl}, DeepEyes~\citep{zheng2025deepeyes}, Mini-o3~\citep{lai2025minio3}, Pixel Reasoner~\citep{su2025pixel}, and Chain-of-Focus~\citep{zhang2025chain}. The most relevant baselines are DeepEyes~\citep{zheng2025deepeyes} and Mini-o3~\citep{lai2025minio3} because they also train multi-turn visual tool use; they therefore isolate the contribution of budget-aware learning more directly than generic VLM baselines.

\noindent\textbf{Metrics.} We report benchmark-standard accuracy. Following the corresponding benchmark protocols, VisualProbe and V\textsuperscript{*} Bench are averaged over 32 repeated runs to reduce stochastic variation, while HR-Bench and MME-RealWorld are evaluated with 8 and 1 run, respectively.

\noindent\textbf{Implementation Details.} Unless otherwise specified, the main results use a Qwen2.5-VL-7B-Instruct backbone~\citep{bai2025qwen25vl} and compare inference under different maximum rollout turns. For more implementation details, including the training hyperparameters and the ablation variants, please refer to Appendix~\ref{sec:appendix-training}.

\begin{figure}[t]
  \centering
  \includegraphics[width=\linewidth]{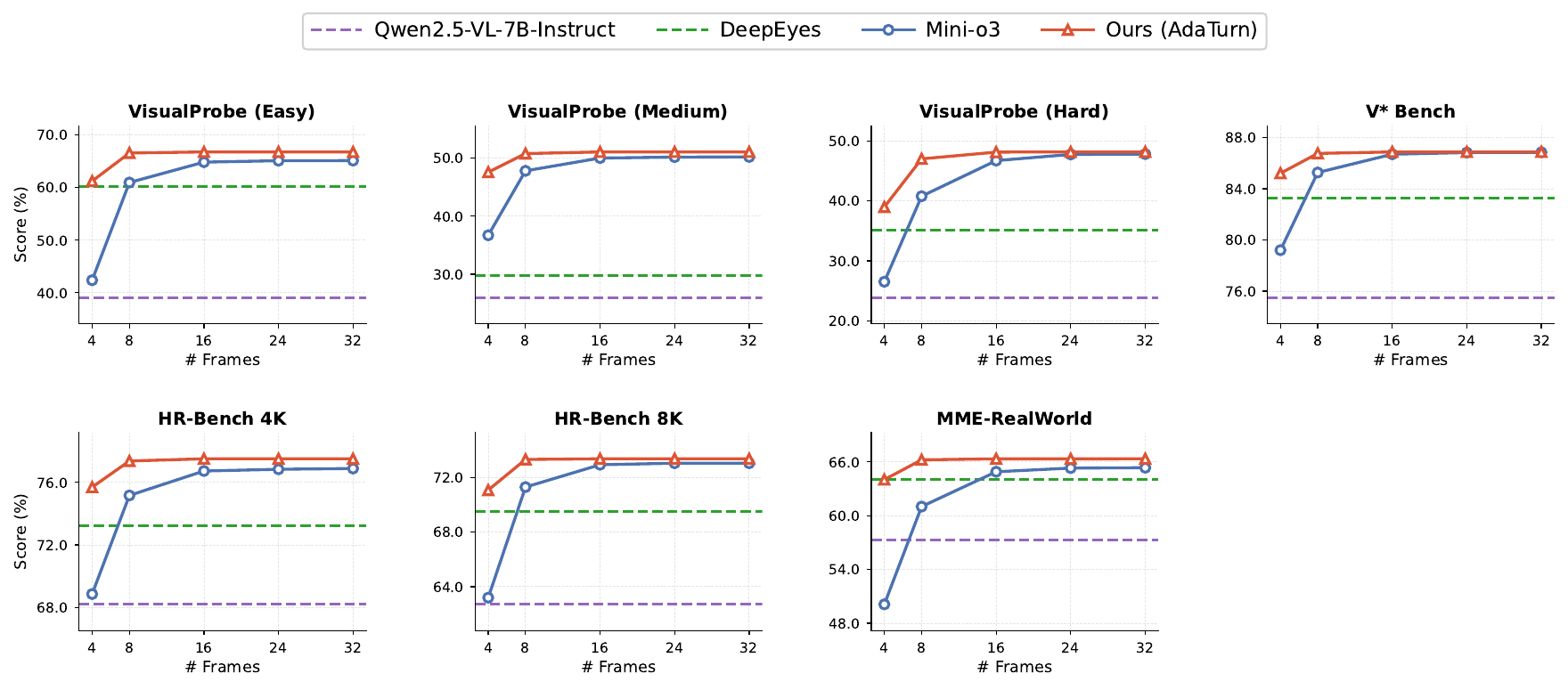}
  \vspace{-2mm}
  \caption{Performance vs. rollout budget. AdaTurn provides a more favorable trade-off than Mini-o3~\citep{lai2025minio3} in the low-budget regime, particularly at four and eight turns, while remaining competitive as additional turns are available.}
  \label{fig:sota}
\end{figure}

\subsection{Quantitative Comparison}
\label{sec:exp-main}
Table~\ref{tab:main} reports the main quantitative comparison. AdaTurn performs particularly well in the low-budget regime, where catastrophic truncation is most consequential. At four turns, AdaTurn improves over Mini-o3~\citep{lai2025minio3} by $+12.5\%$ on VisualProbe-Hard, $+10.9\%$ on VisualProbe-Medium, $+6.0\%$ on V\textsuperscript{*}, $+6.8\%$ on HR-Bench 4K, $+7.9\%$ on HR-Bench 8K, and $+13.9\%$ on MME-RealWorld. These improvements are consistent across VisualProbe~\citep{lai2025minio3}, HR-Bench~\citep{wang2024dc2}, and MME-RealWorld~\citep{zhang2024mmerealworld}, suggesting that the benefit is not confined to a single benchmark family.

At the same time, these gains do not come at the expense of high-budget behavior. At thirty-two turns, AdaTurn matches or exceeds Mini-o3~\citep{lai2025minio3} on VisualProbe-Hard, VisualProbe-Medium, HR-Bench 8K, and MME-RealWorld, while tying on HR-Bench 4K. This indicates that FA-DAPO does not simply encourage earlier termination; rather, it teaches the model to answer appropriately when the budget is limited while still making use of longer rollouts when they are available.

Figure~\ref{fig:sota} presents the same trend from a complementary perspective by plotting performance as a function of the rollout budget. AdaTurn shifts the curve upward in the low-budget region, especially at 4 and 8 turns, while preserving favorable scaling as the budget grows.
Additional rollout visualizations and a failure case analysis are provided in Appendix Sections~\ref{sec:appendix-visualization} and~\ref{sec:appendix-failure}.

\subsection{Ablation Studies}
\label{sec:ablation}

\begin{wrapfigure}[19]{r}{0.40\linewidth}
  \vspace{-1.0em}
  \centering
  \includegraphics[width=\linewidth]{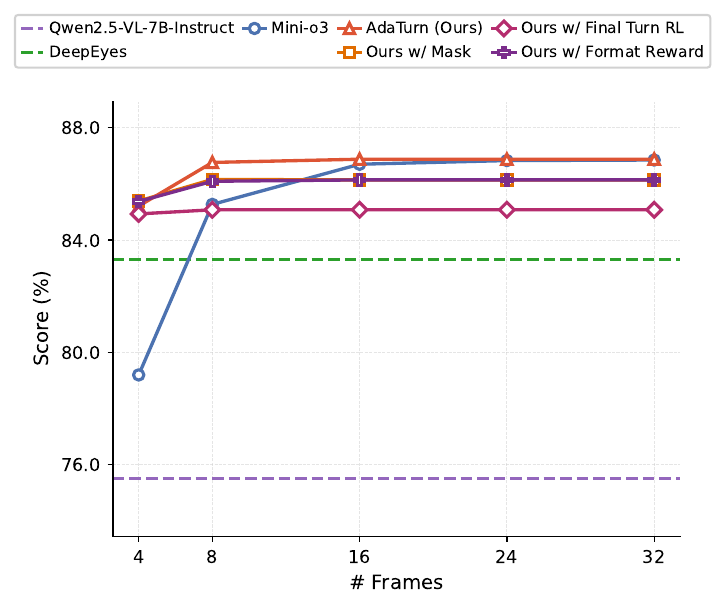}
  \caption{Ablation of the reinforcement learning design. Explicitly training the final-turn answer behavior contributes most of the observed gain. Masking prefix turns, restricting reinforcement learning to only the final forced-answer turn, or adding an extra format reward all underperform the default full-rollout accuracy-only training.}
  \label{fig:ablation}
\end{wrapfigure}

Figure~\ref{fig:ablation} isolates the contribution of the reinforcement learning design. The main takeaway is that forced-answer training is the primary contributor to the improvement: every AdaTurn variant outperforms Mini-o3 at four turns, indicating that explicit supervision of the budget boundary is important for mitigating catastrophic truncation. Among these variants, the default formulation performs best overall, especially once the budget exceeds eight turns.

Appendix~\ref{sec:appendix-ablation} provides the detailed ablation discussion. In brief, masking prefix turns, restricting optimization to only the final forced-answer step, and adding a format reward all underperform the default formulation.

\subsection{Applicability}
\label{sec:applicability}
Table~\ref{tab:backbone} shows that AdaTurn transfers cleanly to other backbone scales. On both Qwen3-VL-4B-Instruct and Qwen3-VL-8B-Instruct~\citep{bai2025qwen3}, AdaTurn outperforms Mini-o3~\citep{lai2025minio3} at four turns across nearly all visual-search benchmarks, and these improvements largely persist at thirty-two turns. This suggests that AdaTurn is not tied to a single base model, but reflects a more general training principle for visual agents.

\begin{table}[t]
  \centering
  \caption{Transfer across backbone scales. AdaTurn improves both Qwen3-VL-4B-Instruct and Qwen3-VL-8B-Instruct under tight budgets and remains competitive when the rollout budget is increased. For VisualProbe and V\textsuperscript{*} Bench, we report Avg@32 to reduce variance caused by randomness. We report Avg@8 and Avg@1 for HR-Bench and MME-RealWorld, respectively.}
  \label{tab:backbone}
  \vspace{0.1cm}
  \tabcolsep=0.25cm
  {
      \begin{threeparttable}
      \resizebox{\linewidth}{!}{
      \begin{tabular}{l|ccc|c|cc|c}
          \toprule
          \multirow{2}{*}{Model}
          & \multicolumn{3}{c|}{VisualProbe}
          & \multirow{2}{*}{V*}
          & \multicolumn{2}{c|}{HR-Bench} & \multirow{2}{*}{MME-RealWorld} \\
          
          \specialrule{0em}{0pt}{1pt}
          \cline{2-4}
          \cline{6-7}
          \specialrule{0em}{0pt}{1pt}
          
          & hard & medium & easy &  & 4K & 8K \\
          \midrule

          Qwen3-VL-4B-Instruct~\citep{bai2025qwen3} & - & - & - & 88.0 & 81.3 & 74.4 & - \\
          + Mini-o3~\citep{lai2025minio3} (4 turns) & 24.7 & 44.3 & 54.9 & 87.0 & 77.2 & 72.4 & 61.2 \\
          + Mini-o3~\citep{lai2025minio3} (32 turns) & 46.3 & 53.9 & 69.1 & 90.2 & \textbf{80.0} & 77.3 & 66.0 \\
          \rowcolor{Gray}+ Ours (4 turns) & \textbf{40.0} & \textbf{51.7} & \textbf{68.4} & \textbf{90.0} & \textbf{79.2} & \textbf{76.6} & \textbf{66.6} \\
          \rowcolor{Gray}+ Ours (32 turns) & \textbf{46.4} & \textbf{54.3} & \textbf{70.6} & \textbf{90.7} & 79.8 & \textbf{78.0} & \textbf{67.1} \\
          
          \specialrule{0em}{0pt}{1pt}
          \hline
          \specialrule{0em}{0pt}{1pt}
          Qwen3-VL-8B-Instruct~\citep{bai2025qwen3} & - & - & - & 90.1 & 82.3 & 78.0 & - \\
          + Mini-o3~\citep{lai2025minio3} (4 turns) & 24.9 & 44.6 & 58.7 & 87.6 & 78.3 & 75.8 & 60.0 \\
          + Mini-o3~\citep{lai2025minio3} (32 turns) & 55.4 & 55.6 & 74.7 & 91.1 & 82.8 & 81.9 & 69.3 \\
          \rowcolor{Gray}+ Ours (4 turns) & \textbf{45.9} & \textbf{54.6} & \textbf{73.8} & \textbf{91.1} & \textbf{82.4} & \textbf{81.4} & \textbf{69.2} \\
          \rowcolor{Gray}+ Ours (32 turns) & \textbf{56.0} & \textbf{57.0} & \textbf{74.8} & \textbf{91.6} & \textbf{83.0} & \textbf{82.3} & \textbf{69.8} \\
          \bottomrule
      \end{tabular}
      }
      \end{threeparttable}
  }
\end{table}

General multimodal results are deferred to Appendix~\ref{sec:appendix-general}. They show that AdaTurn preserves broad OCR and reasoning ability relative to the underlying Qwen2.5-VL-7B-Instruct backbone~\citep{bai2025qwen25vl} while remaining competitive with Mini-o3~\citep{lai2025minio3}.

\section{Conclusion}
\label{sec:conclusion}
We presented AdaTurn, a budget-aware framework for active visual agents under constrained test-time interaction. AdaTurn treats the rollout budget as part of the problem specification, introduces Forced-Answer DAPO to optimize the final-turn boundary explicitly, and combines dynamic-budget training with load-balanced rollout scheduling. Together, these components address catastrophic truncation while preserving the gains of longer rollouts. Empirically, AdaTurn performs favorably in the low-budget regime, remains competitive at larger budgets, transfers to multiple backbone sizes, and preserves broad multimodal capability. More broadly, our results suggest that the next stage of test-time scaling for visual agents is not only about allowing more computation, but about teaching the policy how to reason intelligently under the computation it is actually given.
Limitations and broader impact are discussed in Appendix~\ref{sec:appendix-limitations-impact}.

\bibliographystyle{plainnat}
\bibliography{neurips_2026}

@article{liu2023llava,
  title={Visual instruction tuning},
  author={Liu, Haotian and Li, Chunyuan and Wu, Qingyang and Lee, Yong Jae},
  journal={Advances in Neural Information Processing Systems},
  volume={36},
  year={2024}
}

@article{xu2024llavauhdlmmperceivingaspect,
  title={{LLaVA-UHD}: an {LMM} perceiving any aspect ratio and high-resolution images},
  author={Xu, Ruyi and Yao, Yuan and Guo, Zonghao and Cui, Junbo and Ni, Zanlin and Ge, Chunjiang and Chua, Tat-Seng and Liu, Zhiyuan and Sun, Maosong and Huang, Gao},
  journal={arXiv preprint arXiv:2403.11703},
  year={2024}
}

@article{bai2025qwen25vl,
  title={Qwen2.5-{VL} technical report},
  author={Bai, Shuai and Chen, Keqin and Liu, Xuejing and Wang, Jialin and Ge, Wenbin and Song, Sibo and Dang, Kai and Wang, Peng and Wang, Shijie and Tang, Jun and others},
  journal={arXiv preprint arXiv:2502.13923},
  year={2025}
}

@article{chen2024expanding,
  title={Expanding performance boundaries of open-source multimodal models with model, data, and test-time scaling},
  author={Chen, Zhe and Wang, Weiyun and Cao, Yue and Liu, Yangzhou and Gao, Zhangwei and Cui, Erfei and Zhu, Jinguo and Ye, Shenglong and Tian, Hao and Liu, Zhaoyang and others},
  journal={arXiv preprint arXiv:2412.05271},
  year={2024}
}

@inproceedings{wu2024vstar,
  title={V*: Guided visual search as a core mechanism in multimodal {LLMs}},
  author={Wu, Penghao and Xie, Saining},
  booktitle={Proceedings of the IEEE/CVF Conference on Computer Vision and Pattern Recognition},
  year={2024}
}

@article{wang2024dc2,
  title={Divide, conquer and combine: A training-free framework for high-resolution image perception in multimodal large language models},
  author={Wang, Wenbin and Ding, Liang and Zeng, Minyan and Zhou, Xiabin and Shen, Li and Luo, Yong and Tao, Dacheng},
  journal={arXiv preprint arXiv:2408.15556},
  year={2024}
}

@article{zhang2025chainoffocus,
  title={Adaptive Chain-of-Focus reasoning via dynamic visual search and zooming for efficient {VLMs}},
  author={Zhang, Xintong and Gao, Zhi and Zhang, Bofei and Li, Pengxiang and Zhang, Xiaowen and Liu, Yang and Yuan, Tao and Wu, Yuwei and Jia, Yunde and Zhu, Song-Chun and Li, Qing},
  journal={arXiv preprint arXiv:2505.15436},
  year={2025}
}

@article{zhang2024mmerealworld,
  title={{MME-RealWorld}: Could your multimodal {LLM} challenge high-resolution real-world scenarios that are difficult for humans?},
  author={Zhang, Yi-Fan and Zhang, Huanyu and Tian, Haochen and Fu, Chaoyou and Zhang, Shuangqing and Wu, Junfei and Li, Feng and Wang, Kun and Wen, Qingsong and Zhang, Zhang and others},
  journal={arXiv preprint arXiv:2408.13257},
  year={2024}
}

@article{lai2025minio3,
  title={Mini-o3: Scaling up reasoning patterns and interaction turns for visual search},
  author={Lai, Xin and Li, Junyi and Li, Wei and Liu, Tao and Li, Tianjian and Zhao, Hengshuang},
  journal={arXiv preprint arXiv:2509.07969},
  year={2025}
}

@article{zheng2025deepeyes,
  title={{DeepEyes}: Incentivizing ``thinking with images'' via reinforcement learning},
  author={Zheng, Ziwei and Yang, Michael and Hong, Jack and Zhao, Chenxiao and Xu, Guohai and Yang, Le and Shen, Chao and Yu, Xing},
  journal={arXiv preprint arXiv:2505.14362},
  year={2025}
}

@article{hong2025deepeyesv2,
  title={{DeepEyesV2}: Toward agentic multimodal model},
  author={Hong, Jack and Zhao, Chenxiao and Zhu, ChengLin and Lu, Weiheng and Xu, Guohai and Yu, Xing},
  journal={arXiv preprint arXiv:2511.05271},
  year={2025}
}

@article{lu2025vista,
  title={Scaling agentic reinforcement learning for tool-integrated reasoning in {VLMs}},
  author={Lu, Meng and Xu, Ran and Fang, Yi and Zhang, Wenxuan and Yu, Yue and Srivastava, Gaurav and Zhuang, Yuchen and Elhoseiny, Mohamed and Fleming, Charles and Yang, Carl and others},
  journal={arXiv preprint arXiv:2511.19773},
  year={2025}
}

@article{liu2025agent0vl,
  title={Agent0-{VL}: Exploring self-evolving agent for tool-integrated vision-language reasoning},
  author={Liu, Jiaqi and Xiong, Kaiwen and Xia, Peng and Zhou, Yiyang and Ji, Haonian and Feng, Lu and Han, Siwei and Ding, Mingyu and Yao, Huaxiu},
  journal={arXiv preprint arXiv:2511.19900},
  year={2025}
}

@article{guo2025deepseek,
  title={{DeepSeek-R1}: Incentivizing reasoning capability in {LLMs} via reinforcement learning},
  author={Guo, Daya and Yang, Dejian and Zhang, He and Song, Junxiao and Zhang, Runxin and Xu, Runqi and Zhu, Qihao and Ma, Shirong and Wang, Peiyi and Bi, Xiao and others},
  journal={arXiv preprint arXiv:2501.12948},
  year={2025}
}

@article{shao2024deepseekmath,
  title={Deepseekmath: Pushing the limits of mathematical reasoning in open language models},
  author={Shao, Zhihong and Wang, Peiyi and Zhu, Qihao and Xu, Runxin and Song, Junxiao and Bi, Xiao and Zhang, Haowei and Zhang, Mingchuan and Li, YK and Wu, Yang and others},
  journal={arXiv preprint arXiv:2402.03300},
  year={2024}
}

@article{yu2025dapo,
  title={Dapo: An open-source llm reinforcement learning system at scale},
  author={Yu, Qiying and Zhang, Zheng and Zhu, Ruofei and Yuan, Yufeng and Zuo, Xiaochen and Yue, Yu and Dai, Weinan and Fan, Tiantian and Liu, Gaohong and Liu, Lingjun and others},
  journal={arXiv preprint arXiv:2503.14476},
  year={2025}
}

@article{shen2025vlmr1,
  title={{VLM-R1}: A stable and generalizable {R1}-style large vision-language model},
  author={Shen, Haozhan and Liu, Peng and Li, Jingcheng and Fang, Chunxin and Ma, Yibo and Liao, Jiajia and Shen, Qiaoli and Zhang, Zilun and Zhao, Kangjia and Zhang, Qianqian and Xu, Ruochen and Zhao, Tiancheng},
  journal={arXiv preprint arXiv:2504.07615},
  year={2025}
}

@article{zhang2025r1vl,
  title={{R1-VL}: Learning to reason with multimodal large language models via step-wise group relative policy optimization},
  author={Zhang, Jingyi and Gao, Jiaxing and Pang, Yankai and Zhao, Ruyi and Wang, Xi and Zhang, Jingqun and Hou, Peng and Luo, Rong and Liu, Bin and Huang, Haibin},
  journal={arXiv preprint arXiv:2503.12937},
  year={2025}
}

@article{zhang2025visionaryr1,
  title={Visionary-{R1}: Mitigating shortcuts in visual reasoning with reinforcement learning},
  author={Zhang, Jiaer and Li, Bingqi and Zhou, Wei and Zhao, Rui},
  journal={arXiv preprint arXiv:2505.14677},
  year={2025}
}

@article{snell2024scaling,
  title={Scaling {LLM} test-time compute optimally can be more effective than scaling model parameters},
  author={Snell, Charlie and Lee, Jaehoon and Xu, Kelvin and Kumar, Aviral},
  journal={arXiv preprint arXiv:2408.03314},
  year={2024}
}

@article{openai2024o1,
  title={Learning to reason with {LLMs}},
  author={{OpenAI}},
  journal={OpenAI Blog},
  year={2024},
  url={https://openai.com/index/learning-to-reason-with-llms/}
}

@article{agarwal2025art,
  title={The art of scaling test-time compute for large language models},
  author={Agarwal, Aradhye and Sengupta, Ayan and Chakraborty, Tanmoy},
  journal={arXiv preprint arXiv:2512.02008},
  year={2025}
}

@article{wang2025scaling,
  title={Scaling over scaling: Exploring test-time scaling plateau in large reasoning models},
  author={Wang, Jian and Zhu, Boyan and Leong, Chak Tou and Li, Yongqi and Li, Wenjie},
  journal={arXiv preprint arXiv:2505.20522},
  year={2025}
}

@article{wen2025budgetthinker,
  title={{BudgetThinker}: Empowering budget-aware {LLM} reasoning with control tokens},
  author={Wen, Hao and Wu, Xinrui and Sun, Yi and Zhang, Feifei and Chen, Liye and Wang, Jie and Liu, Yunxin and Liu, Yunhao and Zhang, Ya-Qin and Li, Yuanchun},
  journal={arXiv preprint arXiv:2508.17196},
  year={2025}
}

@article{nogueira2025certainty,
  title={Certainty-guided reasoning in large language models: A dynamic thinking budget approach},
  author={Nogueira, Jo\~{a}o Paulo and Sun, Wentao and Silva, Alonso and Zumot, Laith},
  journal={arXiv preprint arXiv:2509.07820},
  year={2025}
}

@article{gpt4o,
  title={Gpt-4o system card},
  author={Hurst, Aaron and Lerer, Adam and Goucher, Adam P and Perelman, Adam and Ramesh, Aditya and Clark, Aidan and Ostrow, AJ and Welihinda, Akila and Hayes, Alan and Radford, Alec and others},
  journal={arXiv preprint arXiv:2410.21276},
  year={2024}
}

@article{li2024llava,
  title={Llava-onevision: Easy visual task transfer},
  author={Li, Bo and Zhang, Yuanhan and Guo, Dong and Zhang, Renrui and Li, Feng and Zhang, Hao and Zhang, Kaichen and Zhang, Peiyuan and Li, Yanwei and Liu, Ziwei and others},
  journal={arXiv preprint arXiv:2408.03326},
  year={2024}
}

@article{bai2025qwen3,
  title={Qwen3-vl technical report},
  author={Bai, Shuai and Cai, Yuxuan and Chen, Ruizhe and Chen, Keqin and Chen, Xionghui and Cheng, Zesen and Deng, Lianghao and Ding, Wei and Gao, Chang and Ge, Chunjiang and others},
  journal={arXiv preprint arXiv:2511.21631},
  year={2025}
}

@inproceedings{wu2024v,
  title={V?: Guided visual search as a core mechanism in multimodal llms},
  author={Wu, Penghao and Xie, Saining},
  booktitle={Proceedings of the IEEE/CVF Conference on Computer Vision and Pattern Recognition},
  pages={13084--13094},
  year={2024}
}

@inproceedings{li2025dyfo,
  title={Dyfo: A training-free dynamic focus visual search for enhancing lmms in fine-grained visual understanding},
  author={Li, Geng and Xu, Jinglin and Zhao, Yunzhen and Peng, Yuxin},
  booktitle={Proceedings of the Computer Vision and Pattern Recognition Conference},
  pages={9098--9108},
  year={2025}
}

@article{zhang2025chain,
  title={Adaptive Chain-of-Focus Reasoning via Dynamic Visual Search and Zooming for Efficient VLMs},
  author={Zhang, Xintong and Gao, Zhi and Zhang, Bofei and Li, Pengxiang and Zhang, Xiaowen and Liu, Yang and Yuan, Tao and Wu, Yuwei and Jia, Yunde and Zhu, Song-Chun and others},
  journal={arXiv preprint arXiv:2505.15436},
  year={2025}
}

@article{su2025pixel,
  title={Pixel reasoner: Incentivizing pixel-space reasoning with curiosity-driven reinforcement learning},
  author={Wang, Haozhe and Su, Alex and Ren, Weiming and Lin, Fangzhen and Chen, Wenhu},
  journal={arXiv preprint arXiv:2505.15966},
  year={2025}
}

@inproceedings{zheng2024llamafactory,
  title={LlamaFactory: Unified Efficient Fine-Tuning of 100+ Language Models},
  author={Yaowei Zheng and Richong Zhang and Junhao Zhang and Yanhan Ye and Zheyan Luo and Zhangchi Feng and Yongqiang Ma},
  booktitle={Proceedings of the 62nd Annual Meeting of the Association for Computational Linguistics (Volume 3: System Demonstrations)},
  address={Bangkok, Thailand},
  publisher={Association for Computational Linguistics},
  year={2024},
  url={http://arxiv.org/abs/2403.13372}
}

@article{sheng2024hybridflow,
  title   = {HybridFlow: A Flexible and Efficient RLHF Framework},
  author  = {Guangming Sheng and Chi Zhang and Zilingfeng Ye and Xibin Wu and Wang Zhang and Ru Zhang and Yanghua Peng and Haibin Lin and Chuan Wu},
  year    = {2024},
  journal = {arXiv preprint arXiv: 2409.19256}
}

@article{shen2024zoomeye,
  title={ZoomEye: Enhancing Multimodal LLMs with Human-Like Zooming Capabilities through Tree-Based Image Exploration},
  author={Shen, Haozhan and Zhao, Kangjia and Zhao, Tiancheng and Xu, Ruochen and Zhang, Zilun and Zhu, Mingwei and Yin, Jianwei},
  journal={arXiv preprint arXiv:2411.16044},
  year={2024}
}

@inproceedings{lu2022learn,
  title={Learn to Explain: Multimodal Reasoning via Thought Chains for Science Question Answering},
  author={Lu, Pan and Mishra, Swaroop and Xia, Tony and Qiu, Liang and Chang, Kai-Wei and Zhu, Song-Chun and Tafjord, Oyvind and Clark, Peter and Kalyan, Ashwin},
  booktitle={The 36th Conference on Neural Information Processing Systems},
  year={2022}
}

@article{lu2023mathvista,
  title={MathVista: Evaluating Math Reasoning in Visual Contexts with GPT-4V, Bard, and Other Large Multimodal Models},
  author={Lu, Pan and Bansal, Hritik and Xia, Tony and Liu, Jiacheng and Li, Chunyuan and Hajishirzi, Hannaneh and Cheng, Hao and Chang, Kai-Wei and Galley, Michel and Gao, Jianfeng},
  journal={arXiv preprint arXiv:2310.02255},
  year={2023}
}

@inproceedings{masry2022chartqa,
  author = {Masry, Ahmed and Long, Do Xuan and Tan, Jia Qing and Joty, Shafiq and Hoque, Enamul},
  title = {ChartQA: A benchmark for question answering about charts with visual and logical reasoning},
  booktitle = {Proceedings of the 60th Annual Meeting of the Association for Computational Linguistics},
  year = {2022}
}

@inproceedings{mathew2021docvqa,
  author = {Mathew, Minesh and Karatzas, Dimosthenis and Jawahar, C. V.},
  title = {DocVQA: A dataset for VQA on document images},
  booktitle = {Proceedings of the IEEE/CVF Winter Conference on Applications of Computer Vision},
  year = {2021}
}

@misc{simpletir,
  title={SimpleTIR: End-to-End Reinforcement Learning for Multi-Turn Tool-Integrated Reasoning},
  author={Xue, Zhenghai and Zheng, Longtao and Liu, Qian and Li, Yingru and Zheng, Xiaosen and Ma, Zejun and An, Bo},
  year={2025},
  eprint={2509.02479},
  archivePrefix={arXiv},
  primaryClass={cs.LG},
  url={https://arxiv.org/abs/2509.02479}
}

@misc{gspo,
  title={Group Sequence Policy Optimization},
  author={Zheng, Chujie and Liu, Shixuan and Li, Mingze and Chen, Xiong-Hui and Yu, Bowen and Gao, Chang and Dang, Kai and Liu, Yuqiong and Men, Rui and Yang, An and Zhou, Jingren and Lin, Junyang},
  year={2025},
  eprint={2507.18071},
  archivePrefix={arXiv},
  primaryClass={cs.LG},
  url={https://arxiv.org/abs/2507.18071}
}

@article{liu2024ocrbench,
  title={Ocrbench: on the hidden mystery of ocr in large multimodal models},
  author={Liu, Yuliang and Li, Zhang and Huang, Mingxin and Yang, Biao and Yu, Wenwen and Li, Chunyuan and Yin, Xu-Cheng and Liu, Cheng-Lin and Jin, Lianwen and Bai, Xiang},
  journal={Science China Information Sciences},
  volume={67},
  number={12},
  pages={220102},
  year={2024},
  publisher={Springer}
}

@article{tong2024cambrian,
  title={Cambrian-1: A fully open, vision-centric exploration of multimodal llms},
  author={Tong, Shengbang and Brown, Ellis and Wu, Penghao and Woo, Sanghyun and Middepogu, Manoj and Akula, Sai C and Yang, Jihan and Yang, Shusheng and Iyer, Adithya and Pan, Xichen and others},
  journal={Advances in Neural Information Processing Systems},
  volume={37},
  pages={87310--87356},
  year={2024}
}

\appendix

\newpage
\section{Detailed Training Setup}
\label{sec:appendix-training}
Following Mini-o3 \citep{lai2025minio3}, we first initialize AdaTurn by supervised fine-tuning Qwen2.5-VL-7B-Instruct \citep{bai2025qwen25vl} on the Mini-o3 cold-start dataset\footnote{\url{https://huggingface.co/datasets/Mini-o3/Mini-o3-Coldstart-Dataset}}. We then perform reinforcement learning on DeepEyes\_train\_4K\footnote{\url{https://huggingface.co/datasets/Mini-o3/DeepEyes_train_4K}} and VisualProbe\_train\footnote{\url{https://huggingface.co/datasets/Mini-o3/VisualProbe_train}} with dynamic rollout budgets from 4 to 12 turns. We use LLaMA-Factory \citep{zheng2024llamafactory} for supervised fine-tuning and VERL \citep{sheng2024hybridflow} for reinforcement learning, and we follow the released licenses and terms of use of the frameworks, datasets, and pretrained backbones used in our experiments.

\begin{table}[h]
  \centering
  \caption{Training configuration used in our experiments.}
  \label{tab:train-config}
  \small
  \resizebox{\linewidth}{!}{
  \begin{tabular}{lcc}
    \toprule
    Configuration & Supervised fine-tuning & Reinforcement learning \\
    \midrule
    Framework & LLaMA-Factory & VERL \\
    Training data & Mini-o3 cold-start dataset & DeepEyes\_train\_4K + VisualProbe\_train \\
    Batch size & 64 & 64 \\
    Group size & -- & 16 \\
    Mini-batch size & -- & 32 \\
    Learning rate & $1\times 10^{-5}$ & $1\times 10^{-6}$ \\
    Schedule & 10\% warmup + cosine decay & none \\
    Epochs & 3 & 1 \\
    KL loss & -- & none \\
    Reward weights & -- & accuracy 1.0, format 0.0 \\
    Minimum image pixels & 40,000 & 40,000 \\
    Maximum image pixels & 2,000,000 & 2,000,000 \\
    Dynamic rollout turns & -- & 4 to 12 \\
    Freeze vision encoder & true & false \\
    Freeze projection layer & true & false \\
    Freeze LLM & false & false \\
    Context length & 32,768 & 32,768 \\
    Compute & 1 node with 8 A100 GPUs & 2 nodes with 8 A100 GPUs per node \\
    Wall-clock time & about 2 hours & about 3 days \\
    \bottomrule
  \end{tabular}}
\end{table}

\section{Load-Balanced Rollout Assignment Pseudocode}
\label{sec:appendix-load-balance}
Algorithm~\ref{alg:load-balance} gives the full scheduling routine used by AdaTurn when dynamic rollout budgets are enabled. The key idea is to sort requests by assigned rollout budget and greedily place each request on the rollout engine with the smallest current accumulated load. This simple strategy reduces the idle time induced by heterogeneous rollout lengths and complements the main-text discussion in Section~\ref{sec:dynamic-load-balance}.

\begin{algorithm}[t]
   \SetKwInOut{Input}{Input}
   \SetKwInOut{Output}{Output}
    \caption{Load-balanced rollout assignment}
    \label{alg:load-balance}
    \Input{Batch of prompts $P$, set of rollout servers $\mathcal{S}$, rollout turns $R$}
    \Output{Concatenated generation results $O$}
   
   \BlankLine
   \eIf{LoadBalance is \textbf{False}}{
       Divide $P$ into $k$ equal chunks $P_1, \dots, P_k$ where $k = |\mathcal{S}|$\;
       \For{$i = 1$ \KwTo $k$}{
           $O_i \leftarrow \text{Generate}(P_i)$\;
       }
   }{
       Initialize server loads $L \leftarrow [0, \dots, 0]$ for each $s \in \mathcal{S}$\;
       Initialize server assignments $A \leftarrow [\emptyset, \dots, \emptyset]$\;
       Sort indices $Idx$ of $P$ by $R[idx]$ in descending order\;
       
       \For{$idx \in Idx$}{
           $s^* \leftarrow \arg\min_{s} L[s]$\;
           Append $idx$ to $A[s^*]$\;
           $L[s^*] \leftarrow L[s^*] + R[idx]$\;
       }
       
       \For{server $S_i \in \mathcal{S}$}{
           \If{$A[i]$ is not empty}{
               $O_i \leftarrow \text{Generate}(P[A[i]])$\;
           }
       }
   }
    Wait for all $O_i$ and reorder results to match original indices of $P$\;
    \Return $\text{Concat}(O)$
\end{algorithm}

\section{Detailed Ablation Discussion}
\label{sec:appendix-ablation}
Figure~\ref{fig:ablation} shows that explicit supervision at the budget boundary is the main source of improvement. Masking earlier prefix turns degrades performance, which suggests that the final forced-answer decision is best learned together with the preceding search trajectory rather than in isolation. Restricting reinforcement learning to only the final forced-answer turn performs worse still, indicating that budget-aware behavior is a trajectory-level property rather than a single-step correction.

Adding an explicit format reward is also mildly harmful. In our setting, answer extraction already imposes the required structural constraint, so allocating reward mass to formatting does not improve supervision and can distract from the accuracy objective. Taken together, these results support the default full-rollout, accuracy-only FA-DAPO design.

The systems-side ablation in Figure~\ref{fig:load_balance_result} leads to a similar conclusion. The left subfigure reports the minimum and maximum rollout time across engines together with their induced range, while the right subfigure summarizes the gap between the most and least loaded engines. Dynamic rollout budgets create substantial worker imbalance under naive assignment, but the proposed scheduler reduces both statistics and yields an average $1.34\times$ rollout speedup, making budget-aware training materially more efficient in practice.

\section{General Multimodal Capability}
\label{sec:appendix-general}
Table~\ref{tab:general} examines whether AdaTurn preserves general multimodal ability beyond the high-resolution visual-search benchmarks used in the main study. Relative to the base Qwen2.5-VL-7B-Instruct model~\citep{bai2025qwen25vl}, AdaTurn improves OCRBench~\citep{liu2024ocrbench}, ChartQA~\citep{masry2022chartqa}, and CV-Bench~\citep{tong2024cambrian} while remaining competitive on DocVQA~\citep{mathew2021docvqa}, MathVista~\citep{lu2023mathvista}, and ScienceQA~\citep{lu2022learn}. Relative to Mini-o3~\citep{lai2025minio3}, it performs better on OCRBench~\citep{liu2024ocrbench}, ChartQA~\citep{masry2022chartqa}, and CV-Bench~\citep{tong2024cambrian}. These results suggest that budget-aware agent training does not merely specialize the model to a narrow tool-use benchmark, and can be compatible with broad multimodal competence.

\begin{table}[t]
  \centering
  \caption{General multimodal capability after AdaTurn training. Compared with the base Qwen2.5-VL-7B-Instruct model, AdaTurn preserves or improves performance on broad OCR and reasoning benchmarks, indicating that budget-aware agent training does not come at the cost of general visual-language ability.}
  \label{tab:general}
  \vspace{0.1cm}
  \resizebox{\linewidth}{!}{
    \begin{tabular}{l|ccc|c|cc}
      \toprule
      & \multicolumn{3}{c|}{OCR-Related}
      & General
      & \multicolumn{2}{c}{Reasoning} \\
      \specialrule{0em}{0pt}{1pt}
      \cline{2-4}\cline{5-5}\cline{6-7}
      \specialrule{0em}{0pt}{1pt}
      Model
        & OCRBench & ChartQA & DocVQA (val)
        & CV-Bench
        & MathVista (testmini) & ScienceQA (img) \\
      \midrule
      Qwen2.5-VL-7B-Instruct~\citep{bai2025qwen25vl} & 81.5 & 79.6 & 94.6 & 73.9 & 68.2 & \textbf{89.0} \\
      DeepEyes~\citep{zheng2025deepeyes}        & -    & -    & -    & -    & \textbf{70.1} & -  \\
      Mini-o3~\citep{lai2025minio3}             & 83.8 & 77.4 & \textbf{94.8} & 74.4 & 68.8 & 84.5 \\
      \rowcolor{Gray}Ours & \textbf{85.5} & \textbf{80.0} & 93.0 & \textbf{77.6} & 68.1 & 83.7 \\
      \bottomrule
    \end{tabular}
  }
\end{table}

\section{Performance Judge Prompt}
\label{sec:appendix-judge}
To evaluate whether a model prediction matches the ground-truth answer, we use an external judge model. Specifically, we use Qwen3-32B as a binary evaluator that compares the question, the ground-truth answer, and the model prediction, then returns a score of 1 for correct and 0 for incorrect. The exact system prompt and query prompt are shown below.

\begin{promptbox}{System Prompt}
You are an intelligent chatbot designed for evaluating the correctness of generative outputs for question-answer pairs.
Your task is to compare the predicted answer with the correct answer and determine if they match meaningfully. Here's how you can accomplish the task:
------
##INSTRUCTIONS:
- Focus on the meaningful match between the predicted answer and the correct answer.
- Consider synonyms or paraphrases as valid matches.
- Evaluate the correctness of the prediction compared to the answer.
\end{promptbox}

\begin{promptbox}{Query Prompt}
I will give you a question related to an image and the following text as inputs:

1. **Question Related to the Image**: {question}
2. **Ground Truth Answer**: {ground_truth}
3. **Model Predicted Answer**: {prediction}

Your task is to evaluate the model's predicted answer against the ground truth answer, based on the context provided by the question related to the image. Consider the following criteria for evaluation:
- **Relevance**: Does the predicted answer directly address the question posed, considering the information provided by the given question?
- **Accuracy**: Compare the predicted answer to the ground truth answer. You need to evaluate from the following three perspectives:
(1) If the ground truth answer is open-ended, consider whether the prediction accurately reflects the information given in the ground truth without introducing factual inaccuracies. If it does, the prediction should be considered correct.
(2) If the ground truth answer is a definitive answer, strictly compare the model's prediction to the actual answer. Pay attention to unit conversions such as length and angle, etc. As long as the results are consistent, the model's prediction should be deemed correct.
(3) If the model's prediction is irrelevant string, such as punctuation, newlines, tabs, whitespace, or emptry string etc., without a meaningful answer, set the score as 0.
**Output Format**:
Your response should include an integer score indicating the correctness of the prediction: 1 for correct and 0 for incorrect. Note that 1 means the model's prediction strictly aligns with the ground truth, while 0 means it does not.
The format should be "Score: 0 or 1"
\end{promptbox}

\section{AdaTurn Agent Prompts}
\label{sec:appendix-prompts}
AdaTurn uses a small set of structured prompts to control the agent loop. We provide the exact prompts used for the initial system instruction, tool observations, tool errors, and the forced-answer stage when the tool budget is exhausted.

\begin{promptbox}{System Prompt}
You are a helpful assistant. Answer the user's question based on the image provided. Output your thinking process within the <think> and </think> tags. Whenever you find anything unclear, you can zoom in a specific region in the given image to see more clearly by outputing <grounding>{"bbox_2d": [x0, y0, x1, y1], "source": "original_image"}</grounding>, where (x0, y0) and (x1, y1) are the top-left and bottom-right coordinates of the region that you want to zoom in, respectively (suppose the width and height of the image are 1.0), and 'source' refers to the image that you zoom in and could be either 'original_image' or 'observation_i'. Once the final answer is confirmed, put it within <answer> and </answer>.
\end{promptbox}

\begin{promptbox}{Tool Observation Response}
After the above Action {action_turn}, here is the the zoom-in image (Observation {observation_turn}):
<|vision_start|><|image_pad|><|vision_end|>.
Continue your reasoning process inside <think> and </think>. If needed, you can continue to zoom in on the original image or any of the observations, by outputting <grounding> and </grounding> as before. If the final answer is confirmed, put your final answer inside <answer> and </answer>.
\end{promptbox}

\begin{promptbox}{Tool Error Response}
Please analyze the error information obtained from the function tool and adjust your response. Countinue your reasoning process inside <think> and </think>.
\end{promptbox}

\begin{promptbox}{Force-to-Answer Prompt}
<think>I have exhausted my tool-calling budget and must answer the question using the information I have obtained, even though I may make mistakes due to a lack of data.
\end{promptbox}

\section{Visualization of AdaTurn Rollout Trajectories}
\label{sec:appendix-visualization}
\begin{figure}[t]
  \centering
  \includegraphics[width=\linewidth]{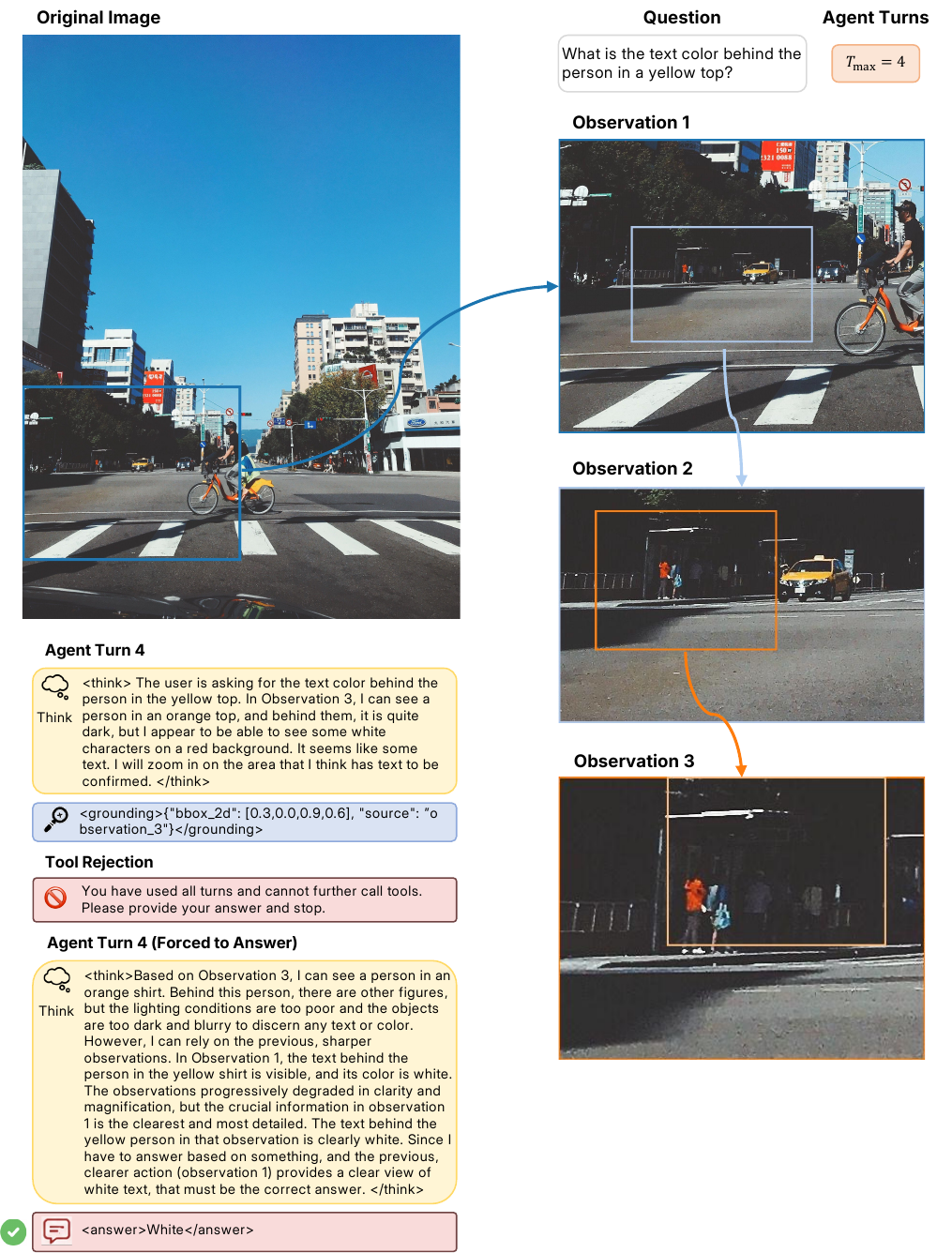}
  \caption{Successful rollout under a four-turn budget. The agent localizes the pedestrian in the yellow top, explores the region behind them, and then answers at the forced-answer step by relying on an earlier, clearer observation that reveals the relevant text color as white. This example illustrates that AdaTurn can retain and use previously collected evidence even when the final crop is less informative.}
  \label{fig:vis1}
\end{figure}

\begin{figure}[t]
  \centering
  \includegraphics[width=\linewidth]{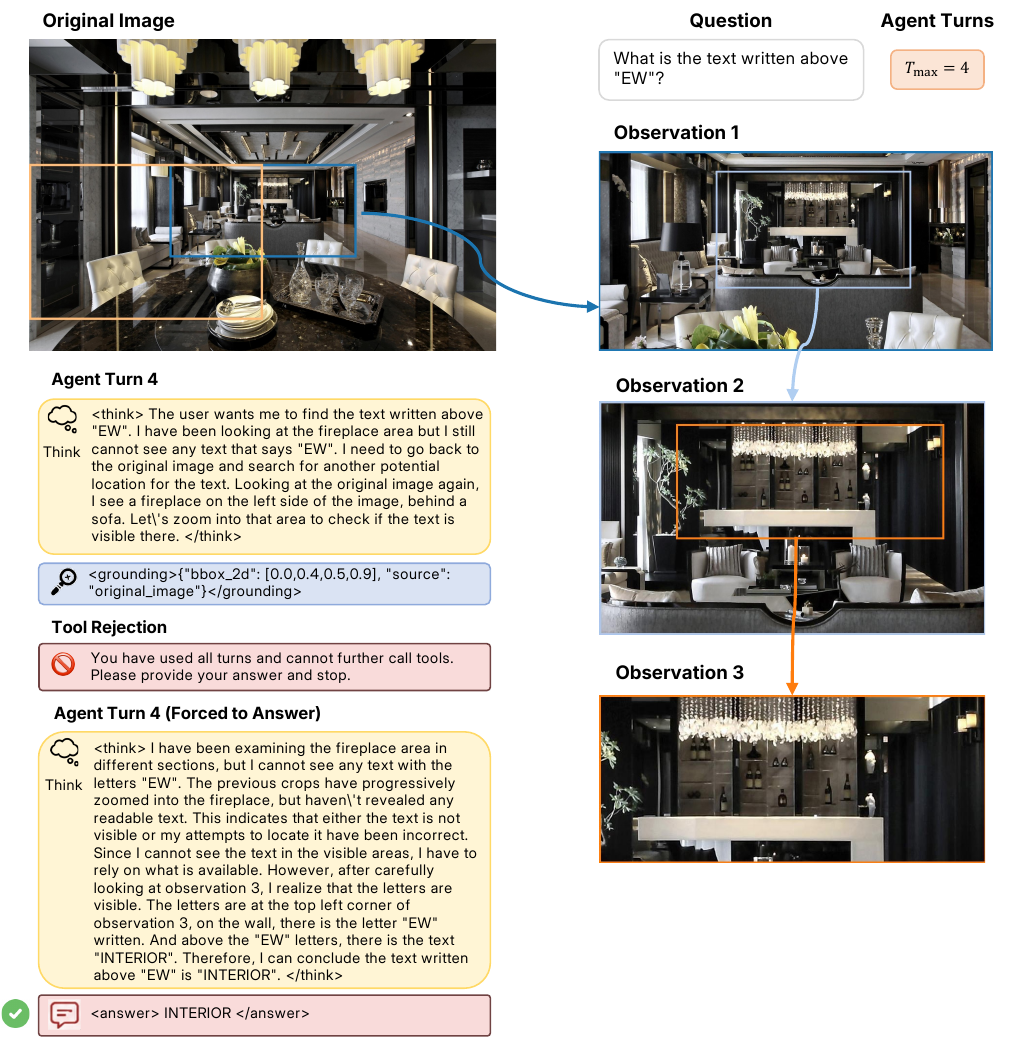}
  \caption{Successful rollout under a four-turn budget. The agent initially searches the fireplace region, fails to confirm the text, then redirects attention to another plausible location and identifies that the text above ``EW'' is ``INTERIOR'' before producing the forced-answer response. This example shows that AdaTurn can revise its search plan rather than committing prematurely to its first hypothesis.}
  \label{fig:vis2}
\end{figure}

\begin{figure}[t]
  \centering
  \includegraphics[width=\linewidth]{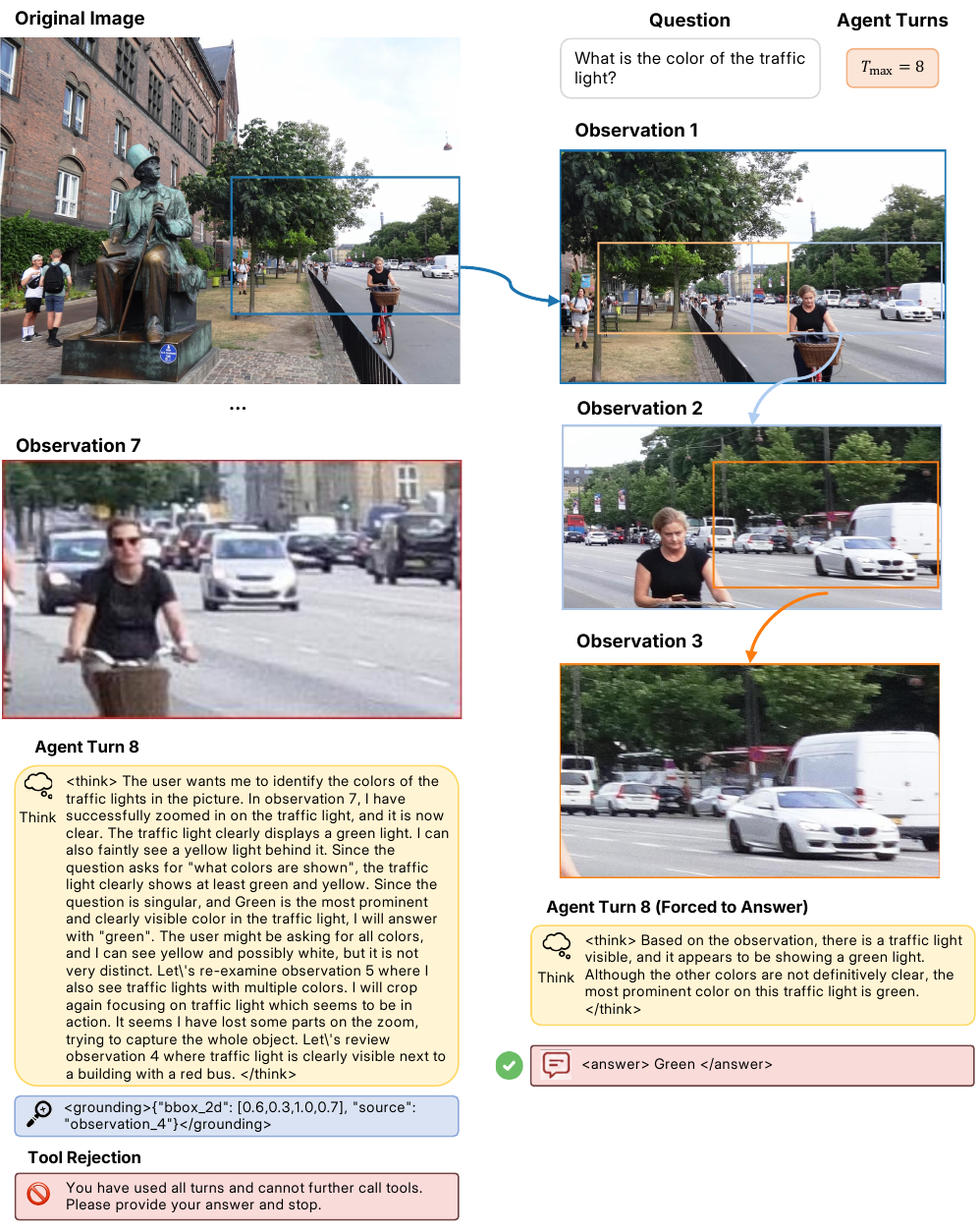}
  \caption{Successful rollout under an eight-turn budget. The agent performs a longer sequence of crops to isolate a distant traffic light and ultimately answers that the light is green. The additional budget enables a deeper search trajectory that would be difficult to complete reliably under a tighter turn limit.}
  \label{fig:vis3}
\end{figure}

\begin{figure}[t]
  \centering
  \includegraphics[width=\linewidth]{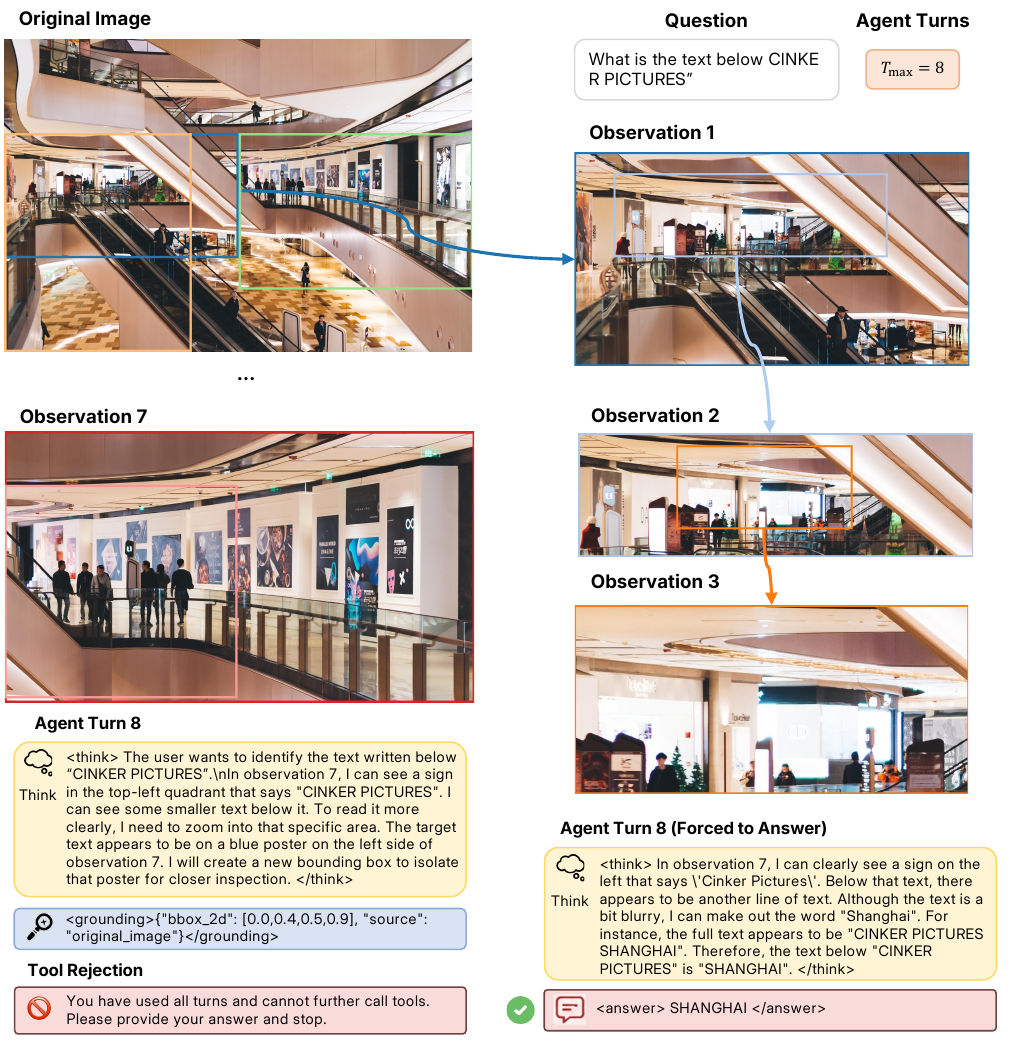}
  \caption{Successful rollout under an eight-turn budget. The agent progressively zooms into a small storefront poster, identifies the sign ``CINKER PICTURES,'' and then uses the final forced-answer step to infer that the text below it is ``SHANGHAI.'' This example illustrates AdaTurn's ability to use longer trajectories for fine-grained text reading while still terminating with a valid answer at the budget boundary.}
  \label{fig:vis4}
\end{figure}

Figures~\ref{fig:vis1} and~\ref{fig:vis2} show representative successful trajectories under a tight four-turn budget, while Figures~\ref{fig:vis3} and~\ref{fig:vis4} show successful trajectories under an eight-turn budget. These examples highlight two qualitative behaviors encouraged by AdaTurn. First, the model does not simply terminate at the budget boundary; instead, it uses the available observations to synthesize a final answer when further grounding is disallowed. Second, the trajectory remains adaptive under different budgets: under $T_{\max}=4$, the model makes compressed decisions based on limited evidence, whereas under $T_{\max}=8$, it can continue refining the crop sequence until the target detail becomes identifiable.

The examples also illustrate that AdaTurn learns a useful form of budget-aware evidence aggregation. In Figure~\ref{fig:vis1}, the final crop is less informative than an earlier observation, but the model still answers correctly by relying on the sharper evidence collected before the final turn. In Figure~\ref{fig:vis2}, the model revisits the original image, shifts attention to a previously overlooked region, and then answers correctly at the forced-answer step. Figures~\ref{fig:vis3} and~\ref{fig:vis4} show that with a larger budget, AdaTurn can sustain longer search trajectories to identify small traffic-light signals and fine-grained scene text, respectively.

\section{Failure Analysis}
\label{sec:appendix-failure}
\begin{figure}[t]
  \centering
  \includegraphics[width=\linewidth]{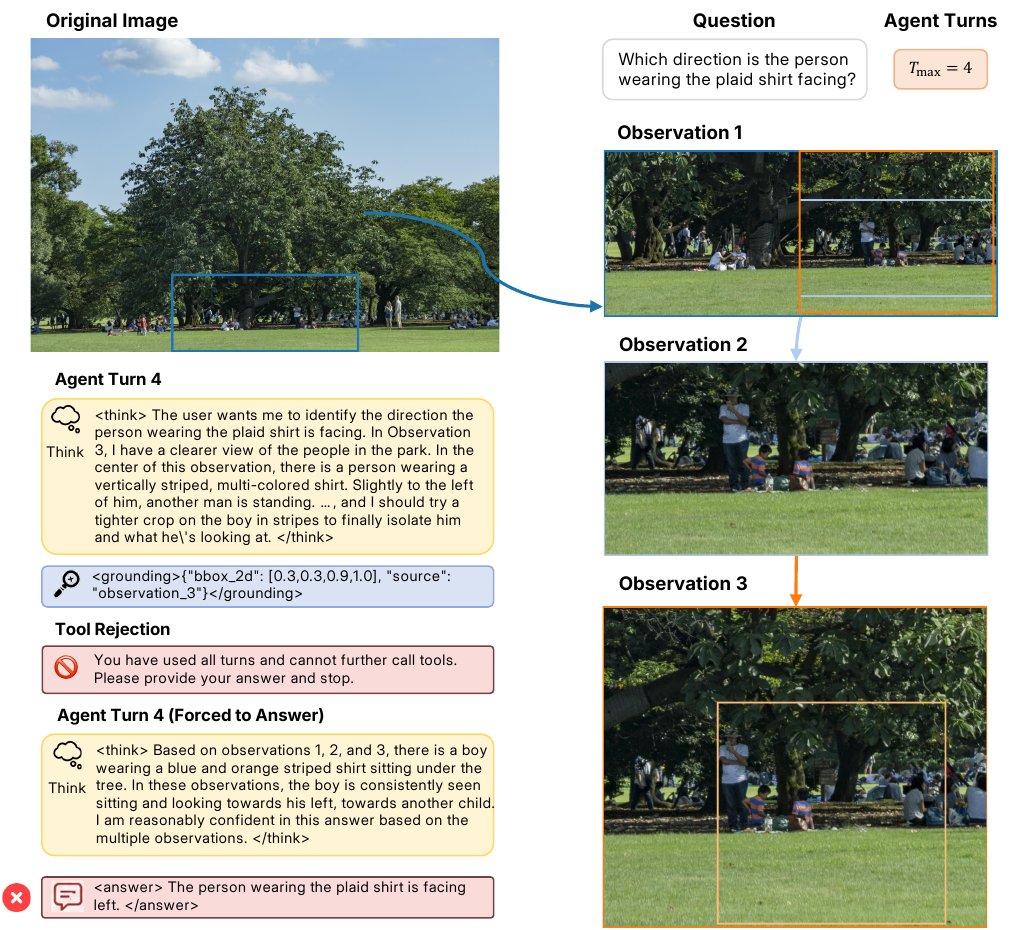}
  \caption{Failure case under a four-turn budget. The agent searches a crowded park scene for the person in a plaid shirt, but the collected crops do not isolate the correct target before the budget is exhausted. At the forced-answer step, the model hallucinates by grounding its decision on a different person and incorrectly answers that the target is facing left.}
  \label{fig:failure1}
\end{figure}

Figure~\ref{fig:failure1} shows a representative failure case under a four-turn budget. The question asks for the direction faced by the person wearing the plaid shirt, but the available observations do not provide sufficient evidence to isolate the correct individual with confidence. After exhausting the tool budget, the model is forced to answer using incomplete information and instead anchors on a different person in the scene, which leads to a hallucinated final answer.

This failure mode is consistent with the intended trade-off of AdaTurn. Forced-answer training mitigates catastrophic truncation by ensuring that the model produces a valid answer even when the search budget is insufficient, but it cannot guarantee correctness when the collected evidence remains ambiguous. In such cases, the model may still hallucinate by overcommitting to an imperfect hypothesis, especially in crowded scenes where small identity distinctions are difficult to resolve within a short rollout.

\section{Limitations and Societal Impact}
\label{sec:appendix-limitations-impact}
\paragraph{Limitations.} Our study is restricted to image-based agents and does not cover video settings, where temporal reasoning and substantially longer interaction horizons introduce additional complexity. We also focus on image crop tools and do not explore broader tool ecosystems such as web search, code execution, or bash interaction.

\paragraph{Societal Impact.} Turn-aware visual agents may benefit latency-sensitive assistive perception, document understanding, and high-resolution inspection systems by producing useful answers under tight compute budgets. At the same time, stronger visual search agents could be misused for surveillance or large-scale automated image inspection, so future releases should follow responsible deployment practices and existing usage restrictions.

\end{document}